\documentclass[11pt,draftcls,onecolumn,journal]{IEEEtran}

\usepackage{amsmath,amsfonts,amssymb,dsfont}
\usepackage{graphicx}
\usepackage{algorithmic}
\usepackage{algorithm}
\usepackage{array} 
\usepackage[caption=false,font=normalsize,labelfont=sf,textfont=sf]{subfig}
\usepackage{textcomp}
\usepackage{stfloats}
\usepackage{url}
\usepackage{verbatim}
\usepackage{cite}
\usepackage[bookmarks,colorlinks]{hyperref}
\usepackage{booktabs}
\usepackage{xcolor}
\usepackage{mathtools}
\usepackage{epsfig}
\usepackage{setspace}
\usepackage{accents}
\usepackage{acronym,flushend} 
\usepackage{enumitem}
\usepackage{xspace}
\usepackage{epstopdf}
\usepackage{tikz}  
\usetikzlibrary{arrows.meta,positioning,shapes.geometric,fit,backgrounds,calc}

\newcommand{\myVec}[1]{{\boldsymbol{#1}}}
\newcommand{\myMat}[1]{{\boldsymbol{#1}}}

\newcommand{\y}{{\myVec{y}}}
\newcommand{\x}{{\myVec{x}}}
\newcommand{\ve}{{\myVec{e}}}
\newcommand{\vw}{{\myVec{w}}}
\newcommand{\vv}{{\myVec{v}}}
\newtheorem{definition}{Definition}

\acrodef{kf}[KF]{Kalman filter}
\acrodef{ekf}[EKF]{extended \ac{kf}}
\acrodef{ai}[AI]{artificial intelligence}
\acrodef{dnn}[DNN]{deep neural network}
\acrodef{nn}[NN]{neural network}
\acrodef{gsp}[GSP]{graph signal processing}
\acrodef{gnn}[GNN]{graph neural network}
\acrodef{rnn}[RNN]{recurrent neural network}
\acrodef{mse}[MSE]{mean-squared error}
\acrodef{snr}[SNR]{signal-to-noise ratio}
\acrodef{ss}[SS]{state-space}
\acrodef{ml}[ML]{machine learning}
\acrodef{mcmc}[MCMC]{Markov chain Monte Carlo}
\acrodef{var}[VAR]{vector autoregressive}
\acrodef{ssm}[SSM]{state-space model}
\acrodef{em}[EM]{expectation-maximization}
\acrodef{pf}[PF]{particle filter}
\acrodef{dag}[DAG]{directed acyclic graph}

\makeatletter\newcommand{\labelname}[1]{\protected@edef\@currentlabelname{#1}}\makeatother

\begin{document}
  
\title{From Granger causality to causal structure in dynamical systems}

\author{V\'ictor Elvira,~\IEEEmembership{Senior Member,~IEEE} \thanks{V.~Elvira is with the Basque Centre for Applied Mathematics (BCAM), Bilbao, Spain, and the School of Mathematics, University of Edinburgh, Edinburgh, UK. e-mail: victor.elvira@ed.ac.uk.}}

%\markboth{IEEE Signal Processing Magazine - Special Issue: From Signals to Causes}%
%{Elvira: From Granger causality to causal structure in dynamical systems}
\markboth{}%
{}
\maketitle

%\section*{}
%\label{sec:abstract}

{\color{black}
\IEEEPARstart{T}{he} idea of testing whether the past of one signal improves the prediction of another can be traced back to Wiener. It was Granger who formalized the concept~\cite{granger1969investigating}. \emph{Granger causality} has become one of the most widely used tools to infer directed interactions between the components of a system. The name however can be misleading: it detects predictive relations among the measured signals, not causal relations among the mechanisms that generate them. This article starts {by} reviewing recent advances in observation-space Granger analysis. It then follows a path to directed structure in latent dynamical models, and asks under which conditions that structure can be read causally.
}

\section*{Introduction}
\label{sec:intro}

Multivariate time series {arise in a wide range of problems in} science and engineering. A recurring question is which components of a system influence the future of others: which brain region drives another, how respiration and arterial pressure influence heart rate~\cite{porta2016wiener}, or which mode of atmospheric variability precedes another across seasons~\cite{elvira2023graphs}. In many applications, only observational data are available (i.e., interventions are not possible), which makes it difficult to draw strong conclusions about the causal mechanisms of the system. Graphical approaches instead infer directed or undirected relations among its components. In signal processing, this line of work is increasingly connected with graph signal processing and topology inference~\cite{ioannidis2019semi,mateos2019connecting}, as well as with the broader causal-inference literature~\cite{runge2023causal}. This tutorial takes a complementary route: it starts from Granger causality and moves the graph from the measured signals to the latent dynamics that generate them.

Granger causality (GC)\cite{granger1969investigating} has been one of the most widely used frameworks for this task for more than half a century. Component $j$ Granger-causes component $i$ if the past of $j$ improves the prediction of $i$ when conditioning on the past of all other components\cite{shojaie2022granger}. Under a vector autoregressive (VAR) model, this corresponds to testing blocks of autoregressive coefficients, whose sparsity pattern defines a directed graph over the observed variables. VAR-based Granger analysis is computationally tractable and widely used across econometrics, neuroscience, and signal processing~\cite{lutkepohl_introduction_2013,shojaie2022granger,barnett2014mvgc}. Significant work has been done recently on high-dimensional VAR models~\cite{basu2015regularized,basu2019low}. 

Observation-space Granger analysis becomes unreliable when the process is only indirectly observed. For instance, temporal filtering, subsampling, and measurement noise in general transform VAR processes into VARMA processes~\cite{barnett2015granger,paulin2026scalable}, while unobserved common drivers induce spurious edges. Nonlinear dynamics introduce additional mismatch, since VAR models capture only linear dependencies. Recent causal-discovery methods such as PCMCI extend observation-space analysis to high-dimensional and nonlinear settings~\cite{runge2019detecting}, but still assume that all relevant drivers are observed. These limitations motivate explicit latent dynamical models.

\begin{figure*}[t]
\centering
\begin{tikzpicture}[
    rung/.style={
        rectangle, rounded corners=3pt, draw, line width=0.8pt,
        text width=3.0cm, minimum height=1.8cm,
        align=center, font=\footnotesize, inner sep=3pt
    },
    rung1/.style={rung, fill=red!7,    draw=red!55},
    rung2/.style={rung, fill=orange!10, draw=orange!60},
    rung3/.style={rung, fill=blue!10,  draw=blue!60},
    rung4/.style={rung, fill=green!12, draw=green!55!black},
    transition/.style={-{Stealth[length=3mm]}, line width=1.2pt, gray!60!black},
    transition_label/.style={font=\scriptsize, align=center, inner sep=2pt}
]
\node[rung1] (r1) at (0,    0) {\textbf{Observation-space Granger graph $\mathcal{G}_{\y}$}\\[1pt]\scriptsize $j\!\to\!i$ iff $\myMat{\Phi}_k(i,j)\!\neq\!0$ in VAR on $\y_t$};
\node[rung2] (r2) at (4.0,  0) {\textbf{Latent directed graph $\mathcal{G}_{\x}$}\\[1pt]\scriptsize {$j\!\to\!i$ iff $j$ directly enters the dynamics of $i$}};
\node[rung3] (r3) at (8.0,  0) {\textbf{Identifiable latent graph $\mathcal{G}_{\x}$}\\[1pt]\scriptsize identifiable transition structure};
\node[rung4] (r4) at (12.0, 0) {\textbf{Causal graph}\\[1pt]\scriptsize structural reading of the transitions};

\draw[transition] (r1) -- (r2);
\draw[transition] (r2) -- (r3);
\draw[transition] (r3) -- (r4);

\node[transition_label, text width=3.4cm] at (2.0,  -1.65) {latent-state modeling};
\foreach \r/\n in {r1/1, r2/2, r3/3, r4/4}
  \node[circle, draw=gray!70!black, fill=white, line width=0.6pt, inner sep=0pt, minimum size=4.2mm, font=\scriptsize\bfseries] at ($(\r.north west)+(0.12,-0.12)$) {\n};
\node[transition_label, text width=3.4cm] at (6.0,  -1.65) {identifiability};
\node[transition_label, text width=3.4cm] at (10.0, -1.65) {structural interpretation}; 
\end{tikzpicture}
\caption{The causal ladder of this tutorial. From left to right: (1) the Granger graph of the observed process; (2) the directed graph of the latent dynamics; (3) the latent graph once it is identifiable; and (4) its causal reading, obtained by interpreting the transition equations as structural mechanisms. The labels below the arrows indicate what is needed to move up one level.} 
\label{fig:causal_ladder}
\end{figure*}
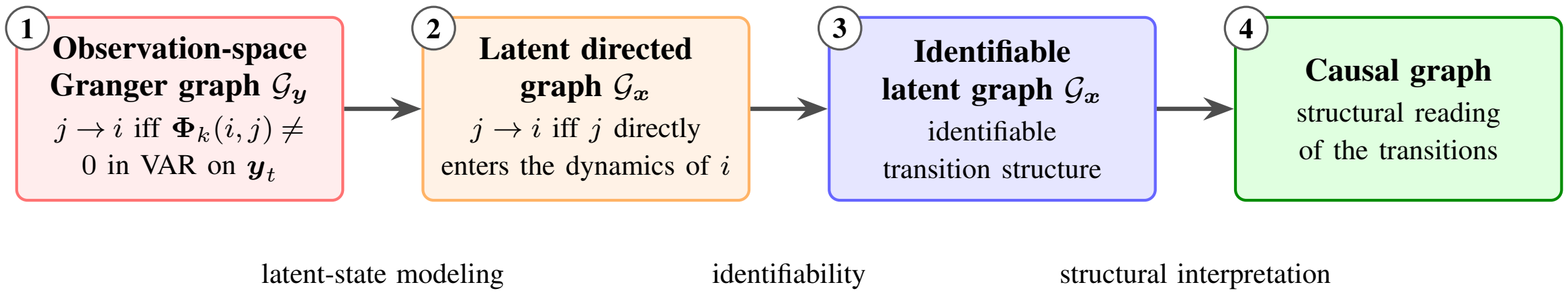

State-space models (SSMs) separate latent dynamics from observations. In a linear Gaussian SSM, a sparse transition matrix $\myMat{A}$ defines a directed graph over latent components, where $A_{ij} \neq 0$ encodes a one-step-ahead influence of state component $j$ on state component $i$. Unlike observation-space Granger estimates, this graph is defined on the latent dynamics, separately from the observation model~\cite{elvira2022graphical,chouzenoux2024sparse}. {This interpretation extends to nonlinear SSMs, where the directed graph describes which latent components enter the transition model of each component.}
In some cases, SSM parameters are identifiable only up to similarity transformations, so interpreting the latent graph requires additional information, such as a known observation model or structural constraints on the model parameters.
The latent transitions can also be given an interventional meaning if they represent structural equations (e.g., known physical unobserved processes). This view is formalized for continuous-time systems by structural dynamical causal models (SDCMs)~\cite{bongers2018causal}. This tutorial follows the resulting \emph{causal ladder} (Figure~\ref{fig:causal_ladder}): from observation-space Granger structure, through latent directed structure, to a causal reading of the latent graph.    

\begin{figure}[t]
\centering
\begin{tikzpicture}[
    xn/.style={circle, draw=blue!60, fill=blue!10, thick,
        minimum size=1.0cm, font=\small},
    yn/.style={circle, draw=green!55!black, fill=green!10, thick,
        minimum size=1.0cm, font=\small},
    arr/.style={-{Stealth[scale=1.1]}, thick},
    obsarr/.style={-{Stealth[scale=1.1]}, thick, black},
    noarr/.style={dashed, thick, gray},
    box/.style={rectangle, rounded corners=4pt,
        draw=gray!85, thick, fill=gray!4},
    blocklabel/.style={font=\footnotesize, black},
    unobslabel/.style={font=\footnotesize\itshape, blue!60},
    obslabel/.style={font=\footnotesize\itshape, green!45!black}
]

\begin{scope}[on background layer]
  \node[box, minimum width=4.4cm, minimum height=2.0cm]
      at (3.3,0.9) {};
  \node[box, minimum width=4.4cm, minimum height=2.0cm]
      at (3.3,-1.8) {};
\end{scope}

\node[xn] (x1) at (0,-0.45) {$x_1$};
\node[above=4pt of x1, font=\footnotesize, black]
    {{respiration}};
\node[below=4pt of x1, unobslabel]
    {(unobserved)};

\node[blocklabel] at (3.3,1.75) {heart rate};
\node[xn] (x2) at (2.2,0.85) {$x_2$};
\node[yn] (y2) at (4.5,0.85) {$y_2$};
\node[unobslabel] at (2.2,0.15) {(unobserved)};
\node[obslabel] at (4.5,0.15) {(observed)};

\node[blocklabel] at (3.3,-0.95) {blood pressure};
\node[xn] (x3) at (2.2,-1.85) {$x_3$};
\node[yn] (y3) at (4.5,-1.85) {$y_3$};
\node[unobslabel] at (2.2,-2.55) {(unobserved)};
\node[obslabel] at (4.5,-2.55) {(observed)};

\draw[arr] (x1) -- (x2);
\draw[arr] (x1) -- (x3);

\draw[noarr] (x2) -- (x3)
    node[midway,right=3pt,font=\footnotesize\itshape]
    {no direct link};

\draw[obsarr] (x2) -- (y2);
\draw[obsarr] (x3) -- (y3);

\end{tikzpicture}
\caption{
Running example: three-node dynamical system used throughout the paper.
The latent respiratory process $x_1$ drives both the future of  heart rate
($x_2$) and blood pressure  ($x_3$). Only $y_2$ and $y_3$ are measured.
Observation-space analysis may therefore infer a spurious directed link
between $y_2$ and $y_3$ due to the shared hidden driver $x_1$,
motivating the transition to latent state-space inference.
}
\label{fig:running_example}
\end{figure}
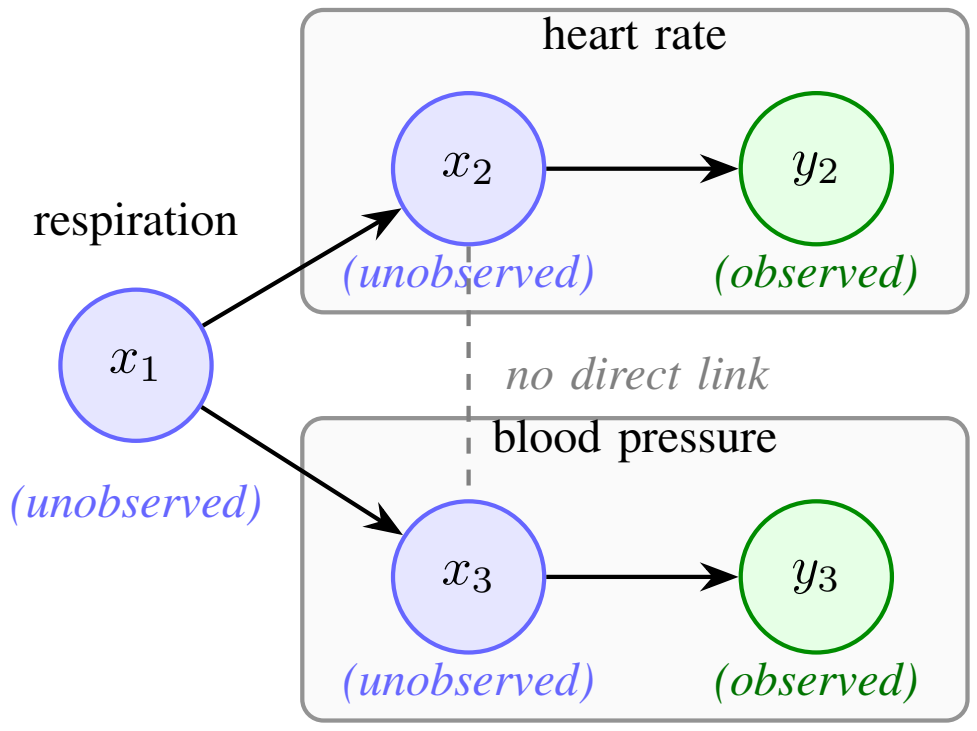

\paragraph{Examples}
A first example considers the cardiovascular setup in Figure~\ref{fig:running_example}, involving three latent physiological processes: $x_1$ (respiratory activity), $x_2$ (heart-rate variability), and $x_3$ (blood-pressure variability). This is a simplified version of the setting in~\cite[p.~303]{porta2016wiener}, where omitting respiration can induce apparent interactions between cardiovascular variables. The true latent structure is
\begin{equation}
x_1 \longrightarrow x_2, \qquad
x_1 \longrightarrow x_3, \qquad
x_2 \not\to x_3, \qquad
x_3 \not\to x_2,
\label{eq:running_example_structure}
\end{equation} 
so that $x_1$ acts as a hidden common driver of both $x_2$ and $x_3$. Observation-space analysis restricted to $(y_2,y_3)$ therefore induces a spurious directed dependence between the observed processes, motivating the transition from Granger analysis in $\y_t$ to latent state-space modeling.

A second example, the five-node chain $x_1\to x_2\to x_3\to x_4\to x_5$, has a different nature: when $x_3$ is unobserved, it is a hidden mediator instead of a hidden common cause, since $x_2$ does affect $x_4$, but only through $x_3$.

\section*{\texorpdfstring{Time-series models and Granger causality}{Time-series models and Granger causality}}
\labelname{Time-series models and Granger causality}\label{sec:background}

\subsection*{Granger causality}
\label{sec:granger_background}

\subsubsection*{Definition}

We consider a stationary stochastic process $\{\y_t\}_{t \geq 1}$, where $\y_t = (y_{t,1}, \ldots, y_{t,n_y})^\top \in \mathbb{R}^{n_y}$; stationarity ensures that the prediction-error variances below do not depend on $t$~\cite{granger1969investigating}. Denote by $\mathcal{H}_{t-1}$ {the information set containing the past of all components up to time $t-1$, and by $\mathcal{H}_{t-1} \setminus y_{\cdot,j}$ the same set except the past of component $j$. The optimal predictor of $y_{t,i}$ given a generic information set $\mathcal{H}$ is denoted $\mathcal{P}(y_{t,i} \mid \mathcal{H})$.

\begin{definition}[Granger causality~\cite{granger1969investigating}]
\label{def:granger_def}
Component $y_{\cdot,j}$ is \emph{Granger causal} for $y_{\cdot,i}$ if
\begin{equation}
  \mathrm{var}\!\left[y_{t,i} - \mathcal{P}\!\left(y_{t,i} \mid \mathcal{H}_{t-1}\right)\right]
  <
  \mathrm{var}\!\left[y_{t,i} - \mathcal{P}\!\left(y_{t,i} \mid \mathcal{H}_{t-1} \setminus y_{\cdot,j}\right)\right].
\label{eq:granger_def}
\end{equation}
\end{definition}

This is a \emph{predictive} notion of directed influence: it does not require, nor directly imply, that $j$ is a cause of $i$ in the interventional sense~\cite{pearl2009causality}.
Definition~\ref{def:granger_def} is the \emph{weak} variant (WLGC), using only past information. The \emph{strong} variant (SLGC)~\cite{caines1976feedback,solo2007causality} extends the information set to contemporaneous variables, capturing instantaneous interactions.  

\subsubsection*{Pairwise, conditional, and block variants}

The definition in~\eqref{eq:granger_def} is \emph{full-conditional}, conditioning on the full past of all components. The \emph{pairwise} variant restricts the information set to the past of $y_{\cdot,i}$ and $y_{\cdot,j}$ alone and is therefore prone to spurious detections (false positives) under hidden common drivers (Figure~\ref{fig:running_example})~\cite{shojaie2022granger}. \emph{Block} variants apply the same comparison to groups of components: a group $\mathcal{A}$ Granger-causes a group $\mathcal{B}$ if removing the past of $\mathcal{A}$ from the full information set increases the prediction error of $\y_{\mathcal{B}}$; in VAR models this is a joint test of zero cross-block coefficients~\cite{lutkepohl_introduction_2013}.

\subsubsection*{The VAR parametric setting}
\label{sec:var_model}

The optimal predictors in~\eqref{eq:granger_def} are conditional expectations, which in practice require a working model to be computable. The most widely used model for GC is the \emph{vector autoregressive} (VAR) model of order $p$:
\begin{equation}
  \y_t = \sum_{k=1}^{p} \myMat{\Phi}_k\, \y_{t-k} + \ve_t,
  \qquad \ve_t \sim \mathcal{N}(\mathbf{0}, \myMat{\Sigma}),
\label{eq:VAR}
\end{equation}
where $\myMat{\Phi}_k \in \mathbb{R}^{n_y \times n_y}$ is the coefficient matrix at lag $k$, and $\ve_t$ is a white-noise innovation with covariance $\myMat{\Sigma}$. Within a correctly specified VAR model, component $j$ Granger-causes component $i$ if
\begin{equation}
  \myMat{\Phi}_k(i,j) \neq 0 \quad \text{for some } k \in \{1, \ldots, p\},
\label{eq:granger_var}
\end{equation}
where $\myMat{\Phi}_k(i,j)$ is the coefficient from component $j$ at lag $k$ to component $i$ at time $t$. The resulting directed graph $\mathcal{G}_{\y}$ has an edge $j\to i$ whenever~\eqref{eq:granger_var} holds. Geweke~\cite{geweke1984measures} introduced a strength measure of Granger causality based on the ratio of prediction error variances:
\begin{equation}
  \mathcal{F}_{j \to i} = \ln \frac{\mathrm{var}[y_{t,i} - \mathcal{P}(y_{t,i} \mid \mathcal{H}_{t-1} \setminus y_{\cdot,j})]}
                                    {\mathrm{var}[y_{t,i} - \mathcal{P}(y_{t,i} \mid \mathcal{H}_{t-1})]}.
\label{eq:geweke}
\end{equation}
$\mathcal{F}_{j \to i} \geq 0$, with $\mathcal{F}_{j \to i} = 0$ if and only if $j$ does not Granger-cause $i$. Thus, Granger causality in the VAR setting reduces to a coefficient-exclusion test, assessed via $F$-tests or likelihood-ratio statistics~\cite{shojaie2022granger,lutkepohl_introduction_2013}.
\subsubsection*{Assumptions and limitations}

The VAR-Granger framework relies on several assumptions:
\begin{itemize}[leftmargin=*]
  \item \textit{Stability}: $\rho(\myMat{\Phi}_{\mathrm{comp}})<1$, where $\rho(\cdot)$ is the spectral radius (maximum eigenvalue modulus) and $\myMat{\Phi}_{\mathrm{comp}}$ the companion matrix of~\eqref{eq:VAR}; this rules out explosive processes.
  \item \textit{Known uniform lag order $p$}: misspecified $p$ biases the graph.
  \item \textit{Causal sufficiency}: all relevant drivers are observed; an unmeasured common driver appears as a spurious Granger link (Figure~\ref{fig:running_example}).
  \item \textit{Perfect observation}: $\y_t$ is free of measurement noise, temporal aggregation, and subsampling.
\end{itemize}
The assumptions are often violated in practice~\cite{shojaie2022granger}, producing spurious or missing Granger links. Hidden variables can affect Granger interpretation in two complementary ways. In the cardiovascular example, $x_2\leftarrow x_1\rightarrow x_3$ with $x_1$ hidden is a \emph{hidden common cause}: intervening on $x_2$ leaves $x_3$ unchanged, yet a Granger link between them may appear. In the five-node chain, a hidden $x_3$ is a \emph{hidden mediator}: intervening on $x_2$ does affect $x_4$, but only through $x_3$, and Granger analysis reports a direct link $x_2\to x_4$; the causal effect is real, and only its reading as a direct edge is spurious.

\subsubsection*{VARMA models}
A vector autoregressive moving-average (VARMA) process of order $(p,q)$ extends the VAR by adding lagged innovations,
\begin{equation}
  \y_t
  =
  \sum_{k=1}^{p}\myMat{\Phi}_k \y_{t-k}
  +
  \ve_t
  +
  \sum_{\ell=1}^{q}\myMat{\Theta}_{\ell}\ve_{t-\ell},
  \qquad
  \ve_t \sim \mathcal{N}(\mathbf{0},\myMat{\Sigma}).
\label{eq:varma}
\end{equation}
VAR models are the subclass with $q=0$, and under standard invertibility conditions a VARMA process admits a VAR representation of infinite order, VAR($\infty$)~\cite[Ch.~11]{lutkepohl_introduction_2013}.

\subsection*{State-space models}
\label{sec:ssm_background}

State-space models (SSMs) describe dynamical systems through a latent state $\x_t = (x_{t,1},\ldots,x_{t,n_x})^\top \in \mathbb{R}^{n_x}$ and an observed signal $\y_t = (y_{t,1},\ldots,y_{t,n_y})^\top \in \mathbb{R}^{n_y}$. In its simplest form, an SSM is specified by a transition density $p(\x_t \mid \x_{t-1})$ and an observation density $p(\y_t \mid \x_t)$, allowing nonlinear, non-Gaussian, and more general dynamical models. We focus first on the linear-Gaussian case,
\begin{align}
  \x_t &= \myMat{A}\,\x_{t-1} + \vw_t, \qquad \vw_t \sim \mathcal{N}(\mathbf{0},\,\myMat{Q}),
  \label{eq:ssm_state} \\
  \y_t &= \myMat{H}\,\x_t + \vv_t, \qquad\quad \vv_t \sim \mathcal{N}(\mathbf{0},\,\myMat{R}),
  \label{eq:ssm_obs}
\end{align}
with initial state $\x_0 \sim \mathcal{N}(\myVec{\mu}_0,\myMat{\Sigma}_0)$ and mutually independent noise sequences. Here $\myMat{A}$ governs the latent dynamics, $\myMat{H}$ maps states to observations, $\myMat{Q}$ is the state-noise covariance, and $\myMat{R}$ is the observation-noise covariance. Filtering, smoothing, and prediction are available in closed form through the Kalman filter and Rauch--Tung--Striebel smoother~\cite{Kalman60,sarkka2023bayesian}.

Any stationary LG-SSM, $\rho(\myMat{A})<1$, induces an observation process that is a linear transformation of a VAR(1) process and admits a VARMA representation~\cite[Sec.~11.6.1]{lutkepohl_introduction_2013}. Conversely, every VARMA process admits an SSM representation, for instance by stacking lags of $\y_t$ and $\ve_t$ into a companion state. Thus, LG-SSMs and VARMA models generate the same class of linear stationary processes, but the parametrizations are not one-to-one~\cite{ljung1999system,lutkepohl_introduction_2013}.

\section*{Observation-space Granger causality in high-dimensional time series}
\label{sec:obs_granger}
In VAR models, the computational and statistical problems arise quickly when the number of series $n_y$ is large ($pn_y^2$ scalar parameters).  Granger causality in a VAR model is encoded in the sparsity pattern of the coefficient matrices: component $j$ Granger-causes component $i$ if and only if $\myMat{\Phi}_k(i,j) \neq 0$ for some $k \in \{1,\ldots,p\}$. Collecting $\{\myMat{\Phi}_k\}_{k=1}^p$ into $\boldsymbol{\Phi} = [\myMat{\Phi}_1 \;\cdots\; \myMat{\Phi}_p]$, the per-lag Granger graphs and the aggregate Granger graph  are
\begin{equation}
  \mathcal{G}_{\y}^{(k)}(i,j) = \mathds{1}\!\left[\myMat{\Phi}_k(i,j) \neq 0\right], \qquad \mathcal{G}_{\y} \;=\; \bigcup_{k=1}^{p}\mathcal{G}_{\y}^{(k)},
\label{eq:granger_graph}
\end{equation}
where the union is taken entrywise.
Estimating $\mathcal{G}_{\y}$ therefore reduces to \emph{support recovery} of $\boldsymbol{\Phi}$~\cite{shojaie2022granger}. This section reviews two families of methods: sparsity-inducing penalties on the coefficients, and strategies that reduce the effective conditioning set. {Both families target the same graph $\mathcal{G}_{\y}$ but bypass the conditional Granger test (one full-conditional regression per pair $(i,j)$), which is computationally and statistically challenging in high dimensions: penalized estimators read off $\mathcal{G}_{\y}$ from the support of a single regularized estimate of $\boldsymbol{\Phi}$, while conditioning-set reduction methods perform explicit independence tests on a data-driven reduced parent set.}

\subsection*{Regularized VAR estimation}
\label{sec:sparse_var}
The methods in this subsection recover $\mathcal{G}_{\y}$ from a single estimate of the VAR coefficients $\boldsymbol{\Phi}$ obtained via penalized loss minimization, rather than through separate pairwise Granger tests.

\subsubsection*{Lasso-penalized VAR}
\label{sec:lasso_var}

 A natural approach to high-dimensional support recovery is to add an $\ell_1$ penalty to the least-squares formulation of the VAR($p$) model. Writing the model in regression form with $\myMat{Y}$ stacking the observations and $\myMat{X}$ the lagged regressors, the \emph{Lasso-VAR} estimator~\cite{basu2015regularized} solves
\begin{equation}
  \widehat{\boldsymbol{\Phi}} = \arg\min_{\boldsymbol{\Phi}}\;
  \frac{1}{T}\|\myMat{Y} - \myMat{X}\boldsymbol{\Phi}^\top\|_F^2
  + \lambda \|\boldsymbol{\Phi}\|_1.
\label{eq:lasso_var}
\end{equation}
{This estimator is formally standard $\ell_1$-penalized regression on a stacked design matrix. However, unlike the classical i.i.d.\ setting, the rows of $\myMat{X}$ are temporally dependent, and the resulting estimation theory depends on spectral properties of the underlying process.} The penalty $\|\boldsymbol{\Phi}\|_1 = \sum_{i,j,k}|\myMat{\Phi}_k(i,j)|$ promotes sparsity in the coefficients.
Under Gaussian innovations with $\myMat{\Sigma}=\sigma^2\myMat{I}$, minimizing the squared-error loss is equivalent to minimizing the negative log-likelihood, so~\eqref{eq:lasso_var} can be read as a penalized maximum-likelihood estimator.
For stable Gaussian VAR processes ($\rho(\boldsymbol{\Phi}_{\mathrm{comp}}) < 1$) with sparse coefficients, Basu \& Michailidis~\cite[Sec.~4]{basu2015regularized} bound the estimation error of the Lasso; the bounds depend on the extremes of the spectral density, so stronger temporal dependence demands more samples.

\subsubsection*{Structured sparsity and latent factor corrections}
\label{sec:structured_var}

A natural refinement under the GC perspective is to enforce sparsity jointly across lags by grouping the coefficients $\{\myMat{\Phi}_1(i,j), \ldots, \myMat{\Phi}_p(i,j)\}$ for each pair $(i,j)$, leading to group-sparse penalties that select or discard entire directed edges across all lags, i.e., promoting sparsity on $\mathcal{G}_{\y}$~\cite{lozano2009grouped,bolstad2011causal}}. Another generalization is {block} Granger causality, where the grouping is over {blocks of dimensions} rather than lags: components are partitioned into groups, and the penalty selects or discards the influence of a whole group on each target component at once~\cite{basu2015network}.

Beyond pair-level sparsity, a few shared latent factors can induce spurious Granger connectivity. Basu et al.~\cite{basu2019low} address this by decomposing the transition matrix of a VAR(1) as $\myMat{\Phi} = \myMat{L}^* + \myMat{R}^*$, with low-rank $\myMat{L}^*$ (capturing latent effects) and sparse $\myMat{R}^*$ (the interpretable Granger structure): the latent effects are accounted for, but not modeled as dynamical processes.

\subsection*{Reducing the conditioning set: domain-guided and data-driven approaches}
\label{sec:pcmci}

Unlike penalized estimators, the methods below perform explicit conditional-independence tests. Each test conditions on a reduced set of variables, chosen from domain knowledge or from the data.
{High-dimensional Granger analysis suffers from two related problems: large conditioning sets reduce statistical power, and strong autocorrelation can make additional predictive contributions from causal parents harder to detect~\cite{runge2019detecting}. The approaches below address this by restricting the conditioning set itself, either through structural constraints or data-driven parent selection.} 

\subsubsection*{Domain-guided conditioning}

Partial knowledge of the physics or topology of the system (e.g., which components can Granger-cause others directly) can be used to decide in advance which components are plausible direct causes. Each conditional Granger test then conditions on a smaller, physically meaningful set.  For instance, in intracavitary electrograms recorded during atrial fibrillation~\cite{luengo2018hierarchical}, electrodes at known positions record the propagation of electrical activation, and electrophysiology rules out certain connections, such as direct links between distant electrodes. Luengo et al.~\cite{luengo2018hierarchical} build the graph as a directed tree: they select as root the electrode with the largest number of significant outgoing pairwise Granger links, and then accept a candidate child of each node only if its Granger causality remains significant given the children already accepted (GS-CaRe); the local variant (LS-CaRe) only considers pairwise-significant links to electrodes within a distance that grows up to a user-defined maximum, linking a distant electrode only if no closer one is connected. Restricting candidates to physically admissible links increases computational efficiency and removes spurious long-range links, at the price that the result is only as good as the prior knowledge it relies on.

\subsubsection*{Data-driven conditioning: PCMCI}
\label{sec:pcmci_alg}

When no structural prior is available, the conditioning sets must be learned from the data. PCMCI~\cite{runge2019detecting} constructs a sparse set of candidate parents for each target and tests directed links within this reduced space. It proceeds in two steps. First (PC step), a condition-selection stage iteratively prunes an initial candidate set using conditional independence tests, yielding a small set $\widehat{\mathcal{P}}(y_{t,i})$ approximating the relevant parents.
Second (MCI step), each candidate link $y_{t-\tau,j} \to y_{t,i}$, {corresponding to the entry $\mathcal{G}_{\y}^{(\tau)}(i,j)$, is tested via the momentary conditional independence (MCI) criterion:
\begin{equation}
  y_{t-\tau,j} \;\perp\!\!\!\perp\; y_{t,i}
  \;\Big|\;
  \widehat{\mathcal{P}}(y_{t,i})\setminus\{y_{t-\tau,j}\},\; \widehat{\mathcal{P}}(y_{t-\tau,j}).
\end{equation}
Conditioning on the parents of the target $y_{t,i}$ isolates the direct effect of the candidate source, while conditioning on the parents of the source $y_{t-\tau,j}$ mitigates inflated test statistics caused by strong autocorrelation \cite{runge2019detecting}. PCMCI supports nonparametric conditional-independence tests, including conditional mutual information, closely related to transfer entropy $T_{j\to i} = I(y_{t,i};\, y_{\cdot,j}^{-} \mid y_{\cdot,i}^{-})$~\cite{schreiber2000measuring}, the information-theoretic analogue of Granger causality (with $T_{j\to i} = \tfrac{1}{2}\mathcal{F}_{j\to i}$ for jointly Gaussian processes~\cite{barnett2009granger}, where $\mathcal{F}_{j\to i}$ is the Geweke measure in~\eqref{eq:geweke}). This captures nonlinear dependencies beyond VAR-based methods. PCMCI$+$~\cite{runge2020discovering} further extends the framework to contemporaneous links ($\tau=0$), analogous to strong Granger causality~\cite{caines1976feedback} in the VAR setting.

\section*{From observation-space Granger to state-space representations}
\label{sec:transition}
 
Vector autoregression is the natural home of Granger causality, but its graphical interpretation relies on strong modeling assumptions. Figure~\ref{fig:failure_modes} summarizes four failure modes motivating the transition to latent state-space models: preprocessing can convert VAR into VARMA, finite-order truncation biases the retained coefficients, VARMA non-causality no longer corresponds to sparsity, and latent confounders induce spurious edges, as in the running example of Figure~\ref{fig:running_example}.

\begin{figure}[t]
\centering
\begin{tikzpicture}[
    cell/.style={
        rectangle, rounded corners=3pt, draw, line width=0.8pt,
        minimum width=7.1cm, minimum height=2.5cm, inner sep=6pt,
        font=\footnotesize
    },
    cellA/.style={cell, fill=red!6,    draw=red!50},
    cellB/.style={cell, fill=green!8,  draw=green!55!black},
    cellC/.style={cell, fill=blue!8,   draw=blue!55},
    cellD/.style={cell, fill=orange!8, draw=orange!60},
    cattag/.style={font=\scriptsize\bfseries, anchor=west},
    title/.style={font=\footnotesize\bfseries},
    desc/.style={font=\scriptsize, text width=6.7cm, align=justify}
]
\node[cellA] (c1) at (0, 0) {};
\node[cattag, red!60!black]   at ($(c1.north west)+(0.15,-0.25)$) {(a) Preprocessing};
\node at ($(c1.center)$) {%
  \begin{tikzpicture}[baseline]
    \node[draw=blue!50, fill=blue!10, rounded corners=2pt, font=\footnotesize, inner sep=3pt] (v) at (0,0) {VAR($p$)};
    \node[draw=orange!60, fill=orange!12, rounded corners=2pt, font=\footnotesize, inner sep=3pt] (vm) at (3.8,0) {VARMA($p',q$)};
    \draw[-{Stealth}, gray!60!black] (v) -- node[above=5pt, font=\scriptsize\itshape]{subsample / filter / +noise} (vm);
  \end{tikzpicture}
};

\node[cellB] (c2) at (7.6, 0) {};
\node[cattag, green!50!black] at ($(c2.north west)+(0.15,-0.25)$) {(b) Model truncation};
\node at ($(c2.center)$) {$\y_t = \sum_{k=1}^{p}\!\myMat{\Phi}_k \y_{t-k} \;+\; \underbrace{\sum_{k=p+1}^{\infty}\!\myMat{B}_k \y_{t-k}}_{\textstyle\color{red!60!black}\text{omitted}} + \myVec{\varepsilon}_t$};

\node[cellC] (c3) at (0, -3.4) {};
\node[cattag, blue!60!black] at ($(c3.north west)+(0.15,-0.25)$) {(c) VARMA opacity};
\node at ($(c3.center)$) {%
  \begin{tikzpicture}[baseline,
      obs/.style={circle, draw=green!60!black, fill=green!10, minimum size=5.5mm, inner sep=0pt, font=\footnotesize},
      tr/.style={-{Stealth[length=2mm]}, line width=0.7pt, gray!60!black}
  ]
    \node[obs] (y1) at (0,0) {$y_1$};
    \node[obs] (y2) at (2,0) {$y_2$};
    \draw[tr, dashed] (y1) -- node[above, font=\normalsize] {?} (y2);
    \node[font=\scriptsize] at (1,-0.7) {$\myMat{\Phi}_{21}=\myMat{\Theta}_{21}=0 \;\not\Leftrightarrow\; y_1 \not\to y_2$};
  \end{tikzpicture}
};

\node[cellD] (c4) at (7.6, -3.4) {};
\node[cattag, orange!70!black] at ($(c4.north west)+(0.15,-0.25)$) {(d) Latent confounder};
\node at ($(c4.center)$) {%
  \begin{tikzpicture}[baseline,
       obs/.style={circle, draw=green!60!black, fill=green!10, minimum size=5.5mm, inner sep=0pt, font=\footnotesize},
       hid/.style={circle, fill=blue!8, draw=blue!55, minimum size=5.5mm, inner sep=0pt, font=\footnotesize, solid},
       tr/.style={-{Stealth[length=2mm]}, line width=0.7pt},
       sp/.style={-{Stealth[length=2mm]}, line width=0.7pt, red!60!black, dashed}
     ]
    \node[hid] (x1) at (1, 0.4) {$x_1$};
    \node[obs] (y2) at (0, -0.5) {$y_2$};
    \node[obs] (y3) at (2, -0.5) {$y_3$};
    \draw[tr] (x1) -- (y2);
    \draw[tr] (x1) -- (y3);
    \draw[sp] (y2) -- (y3);
  \end{tikzpicture}
};
\end{tikzpicture}
\caption{Four failure modes of VAR-based Granger causality. The first three concern the parametrization of the observed process. \emph{(a) Preprocessing}: filtering, subsampling, or measurement noise in general convert a VAR($p$) into a VARMA($p',q$) process. \emph{(b) Model truncation}: finite-order VAR fits truncate the induced VAR$(\infty)$ representation, biasing $\widehat{\myMat{\Phi}}_k$ and distorting $\mathcal{G}_{\y}$. \emph{(c) VARMA opacity}: in VARMA models, Granger non-causality is not encoded as sparsity in $\myMat{\Phi}$ or $\myMat{\Theta}$, but as a rational identity involving both~\cite{boudjellaba1992testing}. The fourth failure mode, \emph{(d) Latent confounding}, is qualitatively different: hidden drivers alter the information set itself, inducing spurious edges in $\mathcal{G}_{\y}$ that observation-space estimators assuming causal sufficiency cannot remove. Together, these effects motivate the transition to latent state-space modeling.}
\label{fig:failure_modes}
\end{figure}
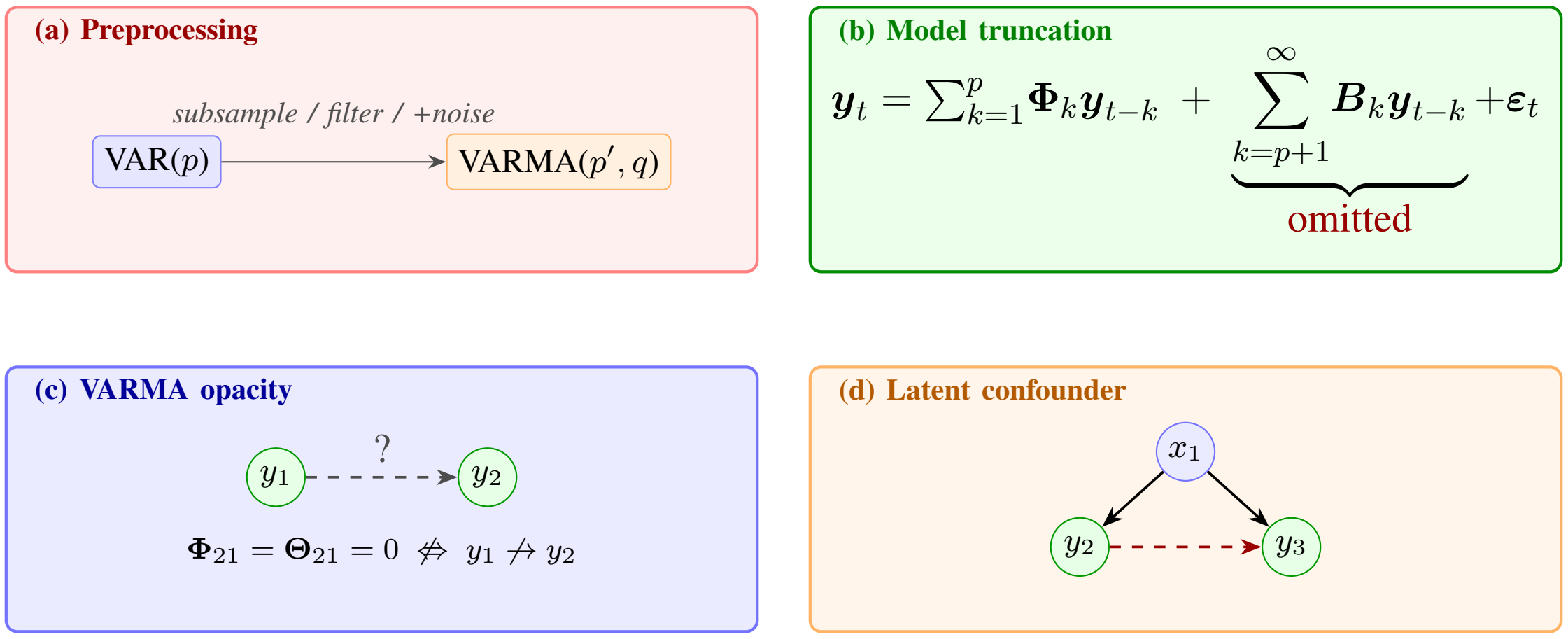
 
\subsection*{\texorpdfstring{Preprocessing and noisy measurements {can} convert VAR to VARMA}{Preprocessing and noisy measurements can convert VAR to VARMA}}
\label{sec:var_breaks}
\label{sec:preprocessing}

Several common preprocessing operations in general turn a VAR into a VARMA, breaking the model assumption of the methods in the previous section (Figure~\ref{fig:var_to_varma}). Observing only a subset of the components of a VAR has the same effect~\cite{barnett2015granger}; we return to it below, since it is the hidden-variable problem.  

\begin{figure}[h]
\centering
\begin{tikzpicture}[
    box/.style={
        rectangle, rounded corners=3pt, draw, line width=0.7pt,
        text width=4.0cm, align=center, font=\footnotesize,
        minimum height=0.7cm, inner sep=2pt
    },
    src/.style={box, fill=blue!10, draw=blue!55},
    op/.style={box, fill=gray!10, draw=gray!55, font=\scriptsize\itshape, text width=2.4cm, minimum height=0.55cm},
    dst/.style={box, fill=orange!15, draw=orange!65},
    arr/.style={-{Stealth[length=2.5mm]}, line width=0.7pt, gray!60!black}
]
\node[src] (var) at (0, 0) {VAR($p$)\\ \mdseries\footnotesize $\y_t = \sum_{k=1}^{p} \myMat{\Phi}_k \y_{t-k} + \ve_t$};
\node[op] (sub)    at (4.4, 1.1)  {subsampling $\y_{mt}$};
\node[op] (filter) at (4.4, 0)    {LTI filter $\sum_k \myMat{G}_k\,\y_{t-k}$};
\node[op] (noise)  at (4.4, -1.1) {additive noise $\y_t + \vv_t$};
\node[dst] (varma) at (9.0, 0) {VARMA($p',q$)\\ \mdseries\scriptsize $\y_t = \sum_{k=1}^{p'} \myMat{\Phi}_k \y_{t-k} + \ve_t + \sum_{k=1}^{q} \myMat{\Theta}_k \ve_{t-k}$};
\draw[arr] (var.east)  to[out=15,  in=180] (sub.west);
\draw[arr] (var.east)  --                  (filter.west);
\draw[arr] (var.east)  to[out=-15, in=180] (noise.west);
\draw[arr] (sub.east)    to[out=0,  in=165] (varma.west);
\draw[arr] (filter.east) --                 (varma.west);
\draw[arr] (noise.east)  to[out=0,  in=195] (varma.west);
\end{tikzpicture}
\caption{Three common operations in general turn a VAR process into a VARMA: temporal subsampling, linear filtering, and additive measurement noise. Any finite-order VAR fit to the transformed series is misspecified.}
\label{fig:var_to_varma}
\end{figure}
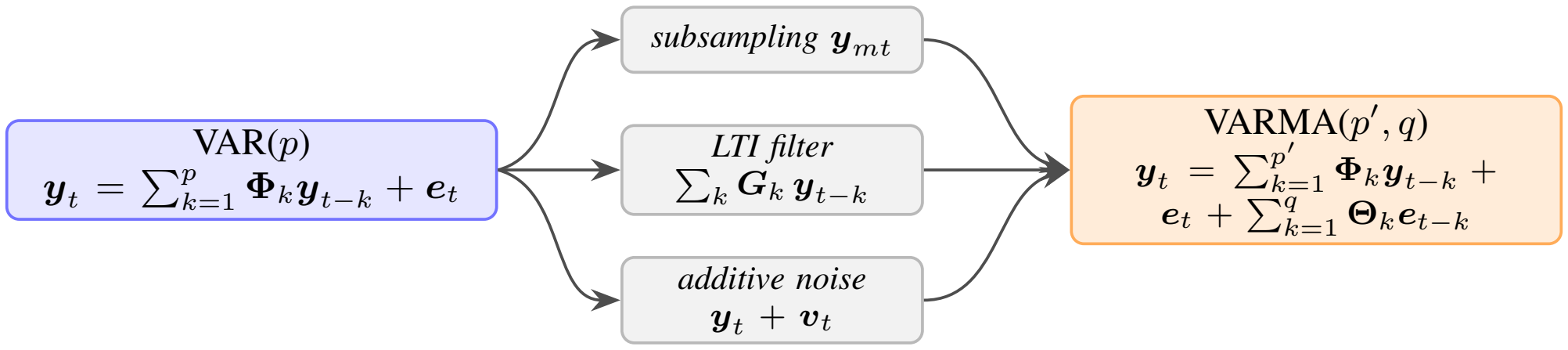

\paragraph{Subsampling}
 
A VAR($p$) process observed at every $m$-th step is in general a VARMA~\cite{solo2007causality}\cite[Sec.~11.6.2]{lutkepohl_introduction_2013}, as an exception, a VAR(1) remains VAR(1), but with transition matrix $\myMat{\Phi}_1^m$, whose graph can differ. Solo~\cite[Sec.~II-B]{solo2007causality} shows that strong Granger causality survives subsampling unless the components become independent, while no such guarantee holds for weak Granger causality. A similar mechanism appears when a continuous-time system is sampled, which matters for the causal reading of the latent graph discussed later.
\paragraph{Filtering}
Linear time-invariant (LTI) filtering can likewise turn a VAR process into a VARMA process~\cite{barnett2015granger}, so that accurately representing the filtered process by a pure VAR can require a substantially higher model order. 
\paragraph{Measurement noise}
If the underlying process evolves as a VAR but is observed with additive independent noise, $\tilde{\y}_t = \y_t + \vv_t$, then $\tilde{\y}_t$ admits a VARMA representation whose MA component is generated by the noise innovation \cite[Sec.~III]{solo2007causality}; see also \cite[Ch.~11]{lutkepohl_introduction_2013}.  

\subsection*{Truncation pathology}
 \label{sec:truncation}

Under standard invertibility conditions, a VARMA process admits a VAR($\infty$) representation, so any finite-order VAR($p$) fit truncates this expansion.  
The omitted high-lag terms carry predictive content that spills into the retained coefficients $\{\myMat{\Phi}_k\}_{k=1}^{p}$, {potentially biasing the Granger graph both ways: low-lag coefficients may absorb high-lag effects (spurious edges), while real influence spread across several lags can drop below detection (missing edges)}~\cite[Ch.~15]{lutkepohl_introduction_2013}. Information criteria~\cite{lutkepohl_introduction_2013} and Bayesian order inference~\cite{binks2024bayesian} can identify a reasonable finite $p$, but residual serial correlation generally remains whenever the true process has an infinite VAR representation. Moreover, the guarantees of the previous section assume a correctly specified finite-order VAR; how sparse VAR graph recovery behaves under VARMA-induced truncation remains much less understood.

\subsection*{Innovations-based Granger and the residual gap}
\label{sec:geweke}
Barnett \& Seth~\cite{barnett2015granger} avoid VAR truncation by representing the observation process with an LG-SSM and computing Granger causality directly from the innovation covariance produced by the Kalman filter. {In steady state, the one-step predictive covariance of $\y_t$ given $\y_{1:t-1}$ is $\myMat{\Sigma} = \myMat{H}\myMat{V}\myMat{H}^\top + \myMat{R}$,} 
where $\myMat{V}$ is the unique stabilizing solution of the discrete algebraic Riccati equation (DARE), not the observation noise $\myMat{R}$. The Geweke measure becomes
\begin{equation}
  \mathcal{F}_{j \to i} = \ln \frac{|\widetilde{\Sigma}_{ii}|}{|\Sigma_{ii}|},
\label{eq:geweke_ssm}
\end{equation}
where $\widetilde{\myMat{\Sigma}}$ is the analogous innovation covariance from the LG-SSM reduced by removing $y_j$ from the observation vector~\cite{geweke1984measures,barnett2015granger}. This computes exact observation-space Granger causality without finite-order VAR approximation. However, the resulting graph is still the observation-space graph $\mathcal{G}_{\y}$, not the latent $\mathcal{G}_{\x}$. The link $(i,j)$ is absent if and only if
\begin{equation}
  \myMat{H}_i\,(\myMat{A} - \myMat{K}\myMat{H})^k\,\myMat{K}_j = \mathbf{0},
  \quad k = 0, 1, \ldots, n_x - 1,
\label{eq:granger_ssm_condition}
\end{equation}
where $\myMat{H}_i$ denotes the $i$-th row of $\myMat{H}$, $\myMat{K}_j$ the $j$-th column of $\myMat{K}$, and $\myMat{K} = \myMat{A}\myMat{V}\myMat{H}^\top\myMat{\Sigma}^{-1}$ is the steady-state Kalman gain in \emph{predictor form} (mapping the innovation into $\widehat{\x}_{t+1|t}$).

\subsection*{An attempt to interpret VARMA models as Granger graphs}
\label{sec:varma_intractable}

 Since preprocessing generally turns the VAR model into a VARMA one, the next question is whether the VARMA parametrization itself admits a transparent GC graph interpretation. Boudjellaba et al.~\cite{boudjellaba1992testing} consider a stationary invertible block-VARMA $\myMat{\Phi}(z)\,\y_t = \myMat{\Theta}(z)\,\ve_t$ partitioned as $\y = (\y_1, \y_2)$, where $\myMat{\Phi}(z)$ and $\myMat{\Theta}(z)$ are the AR and MA matrix polynomials with blocks $\{\myMat{\Phi}_{ij}(z)\}$ and $\{\myMat{\Theta}_{ij}(z)\}$. They show that the Granger non-causality $\y_1 \not\to \y_2$ holds if and only if
  $\myMat{\Phi}_{21}(z) - \myMat{\Theta}_{21}(z)\,\myMat{\Theta}_{11}(z)^{-1}\,\myMat{\Phi}_{11}(z) \;\equiv\; 0$
as a rational identity in $z$. Two consequences follow. First, the AR and MA contributions can {cancel} each other so that $\y_1 \not\to \y_2$ even when both $\myMat{\Phi}_{21}$ and $\myMat{\Theta}_{21}$ are nonzero. Second, an MA cross-block alone can create a Granger link in the absence of any AR cross-coupling. 
Thus, the VARMA parametrization is opaque with respect to the Granger graph: sparsity in either coefficient block is neither necessary nor sufficient for Granger non-causality, precluding any direct reading of the Granger graph from $\myMat{\Phi}$ or $\myMat{\Theta}$ alone.

\subsection*{Latent confounding}
\label{sec:confounding}

Latent confounding is central to causal inference: an unmeasured common driver of two observed components makes any observation-space notion of directed influence ambiguous~\cite{pearl2009causality,spirtes2000causation}. Unlike preprocessing, truncation, or VARMA misspecification, which distort the parametrization, confounding compromises the information set itself. {When a common latent driver acts on multiple observed components, its shared variability can be read by observation-space estimators as direct predictive influence and produce edges that no underlying dynamical mechanism contains.} 

{In causal terminology, hidden variables may act either as confounders (common causes) or as unobserved mediators along causal pathways; throughout this tutorial, both situations are treated under the broader perspective of hidden-variable effects on the observation-space graph. In both cases, the hidden-variable effect can disappear when the omitted process is included explicitly in the latent state, motivating the transition from observation-space analysis on $\y_t$ to latent state-space modeling.}

Suppose that in Figure~\ref{fig:running_example}, $x_2$ and $x_3$ are affected by $x_1$ one step later, and that only $\y_t=(y_{t,2},y_{t,3})^\top$ is observed while $x_{t,1}$ remains hidden. Although no direct interaction exists between $x_{t,2}$ and $x_{t,3}$, the shared influence of $x_{t-1,1}$ may induce predictive dependence between the observed components. Observation-space methods therefore tend to infer a spurious edge between $y_{t,2}$ and $y_{t,3}$. 

This failure concerns methods that assume causal sufficiency, such as Granger causality, Lasso-VAR, and PCMCI; tsFCI~\cite{entner2010causal} and LPCMCI~\cite{gerhardus2020high} drop this assumption and can flag a dependence as possibly due to a hidden common cause, but do not recover the hidden process or its links.

\section*{Learning and interpreting structure in the latent space}
\label{sec:latent_structure}
We now move to latent dynamical models, where the directed graph is defined by the transition structure of the hidden state itself rather than by predictive relations among observations $\y_t$. 
{The central object of this section is the directed graph $\mathcal{G}_{\x}$ defined by the sparsity of the transition kernel of the latent dynamics. We first formalize this graphical interpretation in LG-SSMs and pose the penalized maximum-likelihood estimation problem.} 

\subsection*{Directed structure in latent dynamics}
\label{sec:latent_directed}

In the LG-SSM, the latent state evolves as a VAR(1) on the unobserved process, $\x_t = \myMat{A}\,\x_{t-1} + \vw_t$. The transition matrix $\myMat{A} \in \mathbb{R}^{n_x \times n_x}$ encodes the one-step latent dynamics: entry $A_{ij}$ quantifies the influence of component $j$ at time $t-1$ on component $i$ at time $t$. Equivalently, $A_{ij}\neq 0$ if and only if component $j$ appears in the structural equation of component $i$, $x_{t,i} = \sum_j A_{ij}\,x_{t-1,j} + w_{t,i}$. Its sparsity pattern thus can be read as a \emph{directed graph} $\mathcal{G}_{\x}$ over the $n_x$ latent components \cite{elvira2022graphical}:
\begin{equation}
  j \to i \;\in\; \mathcal{G}_{\x}
  \quad\Longleftrightarrow\quad
  A_{ij} \neq 0.
\label{eq:latent_graph}
\end{equation}
The directed graph $\mathcal{G}_{\x}$ is the latent counterpart of the observation-space Granger graph $\mathcal{G}_{\y}$ in~\eqref{eq:granger_graph}. Under this SSM, the Granger graph on $\y_t$ is a joint function of $\myMat{A}$, $\myMat{H}$, $\myMat{Q}$, and $\myMat{R}$~\eqref{eq:granger_ssm_condition}, mixing latent dynamics with measurement effects.
{Note that higher-order dynamics $\x_t = \sum_{k=1}^{p} \myMat{A}_k\,\x_{t-k} + \vw_t$ can be rewritten as a first-order SSM via companion-form augmentation at the cost of a larger state dimension \cite{chouzenoux2024graphical}.}

\subsection*{Learning sparse latent dynamics}
\label{sec:sparse_latent}

The log-likelihood of $\myMat{A}$ can be computed by marginalizing  over the state trajectory $\x_{1:T}$ as~\cite{sarkka2023bayesian}
\begin{equation}
  \log p(\y_{1:T} \mid \myMat{A})
  = -\sum_{t=1}^{T} \Bigl[
    \tfrac{1}{2}\log\det(2\pi\myMat{S}_t)
    + \tfrac{1}{2}\,\myVec{z}_t^\top \myMat{S}_t^{-1}\myVec{z}_t
  \Bigr],
\label{eq:marginal_ll}
\end{equation}
where $\myVec{z}_t = \y_t - \myMat{H}\myMat{A}\widehat{\x}_{t-1}$ is the {innovation} at time $t$ and $\myMat{S}_t$ is its covariance, both obtained by running the Kalman filter with a given $\myMat{A}$.  
A common approach to recover a sparse $\myMat{A}$ is to solve the {penalized} problem
\begin{equation}
  \widehat{\myMat{A}} = \arg\min_{\myMat{A}}
  \;\underbrace{-\log p(\y_{1:T} \mid \myMat{A})}_{\text{data fit via KF}}
  \;+\; \underbrace{\lambda_A \,\Omega(\myMat{A})}_{\text{structural penalty}},
\label{eq:penalized_ssm}
\end{equation}
where $\Omega$ is a (possibly non-differentiable) convex penalty promoting a desired property of the directed graph $\mathcal{G}_{\x}$: entry-wise sparsity ($\Omega(\myMat{A}) = \|\myMat{A}\|_1$, the default), block-sparsity for pair-level Granger structure, low rank via the nuclear norm, or stability via the operator-norm ball. 
The recursive dependence of $\widehat{\x}_{t-1}$ and $\myMat{S}_t$ on $\myMat{A}$ makes the problem highly nonconvex; although gradients are available, EM-based majorization (GraphEM) and block alternating majorization-minimization (DGLASSO) are used in this case. 

\subsection*{Structure learning in LG-SSMs}
\label{sec:graphem_dglasso}

We now present three methods within the LG-SSM framework: two yield point estimates of the graph as in \cite{elvira2022graphical,chouzenoux2024sparse}, and one takes a fully Bayesian approach that quantifies uncertainty over the graph topology (see also Table~\ref{tab:methods}).  

\subsubsection*{GraphEM (learning the directed graph) \cite{elvira2022graphical}}
\labelname{GraphEM}\label{sec:graphem}

GraphEM computes the MAP estimate of $\myMat{A}$, with the other LG-SSM parameters known, under a penalty $\lambda_A\Omega(\myMat{A})$ (e.g., $\ell_1$ norm). It modifies the EM algorithm for SSMs~\cite{shumway1982approach}: the E-step runs the Kalman filter and RTS smoother with the current $\myMat{A}^{(i)}$ to build a minorizer of the log-likelihood, and the M-step minimizes the penalized surrogate. Under mild conditions, the objective decreases monotonically and every cluster point of the iterates is a critical point~\cite[Thm.~1]{elvira2022graphical}.%

\subsubsection*{DGLASSO (learning directed and undirected graphs)~\cite{chouzenoux2024sparse}}
\labelname{DGLASSO}\label{sec:dglasso}

DGLASSO jointly estimates $\myMat{A}$ and the state-noise precision $\myMat{P}=\myMat{Q}^{-1}$ by minimizing the negative marginal log-likelihood~\eqref{eq:marginal_ll} plus $\lambda_A\|\myMat{A}\|_1+\lambda_P\|\myMat{P}\|_1$, with a block alternating majorization-minimization scheme that converges to a critical point under mild conditions~\cite[Thm.~2]{chouzenoux2024sparse}. As in the graphical Lasso~\cite{friedman2008sparse,lauritzen1996graphical}, the off-diagonal zeros of $\myMat{P}$ define an undirected graph $\mathcal{G}_P$ of conditional dependencies between noise components, so DGLASSO returns a dual description: dynamical influences in $\mathcal{G}_{\x}$ and contemporaneous dependencies in $\mathcal{G}_P$.

{Figure~\ref{fig:dglasso_vs_var} illustrates the parametrization gap with a small simulation. A sparse Markov-1 SSM with $9$ transition entries and $2$ undirected precision edges (panel~(c)) induces in observation space as a much denser VAR representation: $50$ directed OLS-VAR coefficients plus $6$ residual-precision edges (panel~(a)), or $25$ coefficients after $\ell_1$ regularization (panel~(b)). The lag-$2$ and lag-$3$ ``ghost'' edges persist under Lasso because they reflect the effective higher-order dependence induced in observation space by the latent dynamics and measurement noise.}

\begin{figure*}[t]
\centering
\begin{tikzpicture}[
    yn/.style={circle, draw=green!55!black, fill=green!10, minimum size=4mm, inner sep=0pt},
    xn/.style={circle, draw=blue!60, fill=blue!10, minimum size=4mm, inner sep=0pt},
    var_thick/.style={->, line width=0.7pt,  green!45!black, draw opacity=1.00, >=Stealth, shorten >=1pt},
    var_med/.style  ={->, line width=0.42pt, green!45!black, draw opacity=0.70, >=Stealth, shorten >=1pt},
    var_thin/.style ={->, line width=0.25pt, green!45!black, draw opacity=0.32, >=Stealth, shorten >=1pt},
    ssm_edge/.style={->, line width=0.8pt, blue!65!black, >=Stealth, shorten >=1pt},
    self_edge/.style={->, line width=0.4pt, blue!50, >=Stealth, shorten >=1pt},
    p_edge/.style={red!70!black, line width=0.7pt},
    column_label/.style={font=\scriptsize\itshape, anchor=south},
    panel_title/.style={font=\small\bfseries, anchor=south}
]
\begin{scope}[xshift=0cm]
\node[panel_title] at (2.0, 0.55) {(a) OLS VAR($p=3$) fit};
\foreach \col/\xpos/\lab in {1/0/{$\y_{t-3}$}, 2/1.4/{$\y_{t-2}$}, 3/2.8/{$\y_{t-1}$}, 4/4.2/{$\y_{t}$}} {
    \foreach \row in {1,2,3,4,5} {
        \node[yn] (l\col\row) at (\xpos, -\row*0.55) {};
    }
    \node[column_label] at (\xpos, -0.15) {\lab};
}
\foreach \i/\j in {1/1, 1/2, 2/2, 2/3, 3/3, 3/4, 4/4, 4/5, 5/5}
    { \draw[var_thick] (l3\i) -- (l4\j); }
\foreach \i/\j in {1/3, 3/2, 4/3, 2/4}
    { \draw[var_med]   (l3\i) -- (l4\j); }
\foreach \i/\j in {2/1, 3/1, 4/2, 1/4, 5/4, 1/5, 2/5, 3/5}
    { \draw[var_thin]  (l3\i) -- (l4\j); }
\foreach \i/\j in {1/1, 1/2, 2/2, 2/3, 3/3, 3/4, 4/4, 4/5, 5/5}
    { \draw[var_med]   (l2\i) to[bend left=12] (l4\j); }
\foreach \i/\j in {2/1, 1/3, 2/4, 3/5}
    { \draw[var_thin]  (l2\i) to[bend left=12] (l4\j); }
\foreach \i/\j in {1/2, 2/3, 3/4, 4/4, 4/5, 5/5}
    { \draw[var_med]   (l1\i) to[bend left=22] (l4\j); }
\foreach \i/\j in {1/1, 2/1, 3/1, 2/2, 3/2, 1/3, 3/3, 2/4, 5/4, 3/5}
    { \draw[var_thin]  (l1\i) to[bend left=22] (l4\j); }
\foreach \i/\j in {1/2, 2/3, 3/4, 4/5}
    { \draw[p_edge, bend left=20] (l4\i) to (l4\j); }
\foreach \i/\j in {1/3, 2/4}
    { \draw[p_edge, bend left=40] (l4\i) to (l4\j); }
\end{scope}
\begin{scope}[xshift=6cm]
\node[panel_title] at (2.0, 0.55) {(b) Lasso-VAR($p=3$) fit};
\foreach \col/\xpos/\lab in {1/0/{$\y_{t-3}$}, 2/1.4/{$\y_{t-2}$}, 3/2.8/{$\y_{t-1}$}, 4/4.2/{$\y_{t}$}} {
    \foreach \row in {1,2,3,4,5} {
        \node[yn] (m\col\row) at (\xpos, -\row*0.55) {};
    }
    \node[column_label] at (\xpos, -0.15) {\lab};
}
\foreach \i/\j in {1/1, 2/2, 2/3, 3/3, 3/4, 4/4, 4/5, 5/5}
    { \draw[var_thick] (m3\i) -- (m4\j); }
\foreach \i/\j in {1/2, 3/2, 4/3}
    { \draw[var_med]   (m3\i) -- (m4\j); }
\foreach \i/\j in {5/4}
    { \draw[var_thin]  (m3\i) -- (m4\j); }
\foreach \i/\j in {2/2, 2/3, 3/3, 3/4, 4/4, 4/5, 5/5}
    { \draw[var_med]   (m2\i) to[bend left=12] (m4\j); }
\foreach \i/\j in {1/2}
    { \draw[var_thin]  (m2\i) to[bend left=12] (m4\j); }
\foreach \i/\j in {4/5}
    { \draw[var_med]   (m1\i) to[bend left=22] (m4\j); }
\foreach \i/\j in {2/3, 3/4, 4/4, 5/5}
    { \draw[var_thin]  (m1\i) to[bend left=22] (m4\j); }
\end{scope}
\begin{scope}[xshift=12.3cm]
\node[panel_title] at (0.9, 0.55) {(c) Latent SSM (DGLASSO)};
\foreach \col/\xpos/\lab in {1/0/{$\x_{t-1}$}, 2/1.8/{$\x_{t}$}} {
    \foreach \row in {1,2,3,4,5} {
        \node[xn] (s\col\row) at (\xpos, -\row*0.55) {};
    }
    \node[column_label] at (\xpos, -0.15) {\lab};
}
\foreach \i/\j in {1/2, 2/3, 3/4, 4/5} { \draw[ssm_edge] (s1\i) -- (s2\j); }
\foreach \i in {1,2,3,4,5} { \draw[self_edge] (s1\i) -- (s2\i); }
\draw[p_edge, bend left=40] (s21) to (s23);
\draw[p_edge, bend left=40] (s22) to (s24);
\end{scope}
\end{tikzpicture}
\caption{{\textbf{Observation-space sparsity versus latent-space sparsity.}
Data are generated from the sparse LG-SSM shown in panel~(c): a 5-dimensional Markov-1 latent process with sparse transition matrix $\myMat{A}$ (chain $1\!\to\!2\!\to\!3\!\to\!4\!\to\!5$ plus self-loops) and sparse state-noise precision graph $\mathcal{G}_P$ (edges $1\!-\!3$ and $2\!-\!4$), observed through identity $\myMat{H}$ with additive measurement noise. The progression
$56\;\text{(OLS)} \rightarrow 25\;\text{(Lasso)} \rightarrow 9+2\;\text{(latent SSM)}$
illustrates the parametrization gap between observation-space and latent-space representations. We use $T=2\cdot10^5$ samples.
\textbf{(a)}~OLS VAR($p=3$) fit on the observations, together with the residual precision graph. Only arrows arriving at time $t$ are displayed. Because the noisy observation process is VARMA, the induced finite-order VAR representation becomes dense even though the latent dynamics are sparse.
\textbf{(b)}~Row-wise Lasso-VAR($p=3$) on the same data. The $\ell_1$ penalty removes many weak coefficients, but several lag-$2$ and lag-$3$ links persist, reflecting truncation effects and the effective infinite-order dependence induced in observation space.
\textbf{(c)}~DGLASSO~\cite{chouzenoux2024sparse} recovers the true sparse dual graph $(\mathcal{G}_{\x},\mathcal{G}_P)$ in latent space.}}
\label{fig:dglasso_vs_var}
\end{figure*}
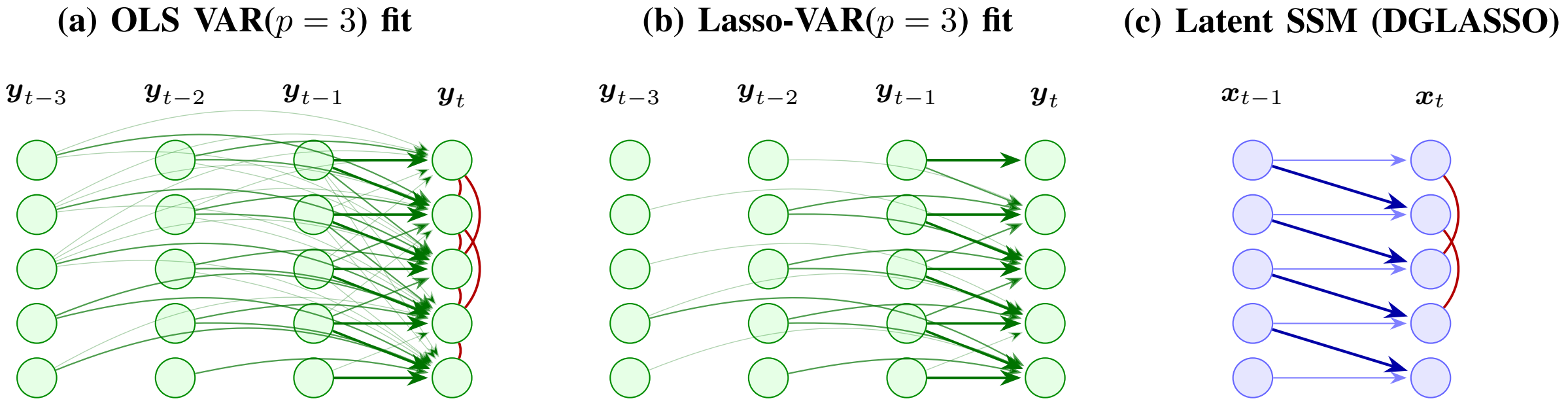

\subsubsection*{Bayesian formulations}
\label{sec:bayesian}
 A complementary perspective is to treat $\myMat{A}$ as a random variable and compute its \emph{posterior distribution} $p(\myMat{A} \mid \y_{1:T})$, quantifying uncertainty over both the edge weights and the graph topology, i.e., also $p(\mathcal{G}_{\x} \mid \y_{1:T})$. SpaRJ~\cite{cox2023sparse} is a fully Bayesian approach for the LG-SSM based on {reversible-jump Markov chain Monte Carlo} (RJMCMC) \cite{green1995reversible}, interpreting the sparsity pattern of $\myMat{A}$, $\mathcal{G}_{\x}$, as a {model indicator}: each possible support defines a different model, and the chain explores model space and parameter space hierarchically. Transition kernels add or remove edges (model jumps) and update the nonzero entries (within-model moves); the Kalman filter provides the marginal likelihood for the acceptance ratio. The output samples can be processed for instance to  estimate the probability $\Pr(A_{ij} \neq 0 \mid \y_{1:T})$ for each directed edge.

In the overall picture, the Bayesian route reports uncertainty over graphs instead of a thresholded estimate; its priors play the role of the structural information (sparsity priors for the $\ell_1$ penalty, zero prior mass for masked edges), and a non-identified model can show up as a posterior spread over the similarity class. Particle MCMC and SMC$^2$ extend it to nonlinear SSMs by replacing the Kalman likelihood with particle-filter estimates~\cite{kantas2015particle}. The cost (one filtering pass per iteration over up to $2^{n_x^2}$ supports) limits it to moderate $n_x$.

\subsection*{Beyond linear-Gaussian dynamics}
\label{sec:graphgrad}

Many systems of scientific interest are however nonlinear (e.g., chaotic physical systems, biological oscillators); forcing a linear model distorts both the inferred dynamics and the recovered graph. Recent approaches in dynamical systems have explored sparse nonlinear representations and polynomial approximations for discovering governing equations from data~\cite{brunton2016discovering}. GraphGrad~\cite{cox2025graphgrad} extends LG-SSMs by replacing linear transitions with sparse polynomial dynamics, which can be interpreted as polynomial approximations of more general transition functions. Note that several canonical dynamical systems of practical interest (Lorenz~63 and~96, Lotka-Volterra predator-prey models, compartmental epidemiological models such as SIR and SEIR) admit exactly polynomial, often quadratic, state dynamics.
 
GraphGrad parametrizes the transition as
\begin{equation}
  \x_t \sim p(\x_t \mid \x_{t-1};\, \myMat{C})
  = \mathcal{N}\!\bigl(
      \myMat{C}\,\myVec{m}(\x_{t-1}; \myMat{D}),
      \myMat{Q}
    \bigr),
\label{eq:graphgrad_model}
\end{equation}
where $\myVec{m}(\x; \myMat{D}) \in \mathbb{R}^M$ is the vector of all monomials of degree at most $d$ in $\x$, whose exponents are encoded in the columns of $\myMat{D} \in \mathbb{N}_0^{n_x \times M}$ (fixed once $n_x$ and $d$ are chosen), and $\myMat{C} \in \mathbb{R}^{n_x \times M}$ is the learnable coefficient matrix~\cite{cox2025graphgrad}.
Explicitly, the $\ell$-th column $\myMat{D}_{:,\ell} = [D_{1\ell}, \ldots, D_{n_x\ell}]^\top$ of $\myMat{D}$ is the exponent vector that defines the $\ell$-th monomial,
\begin{equation}
  \color{black}
  \myVec{m}(\x; \myMat{D}) = \bigl[m_1(\x; \myMat{D}), \ldots, m_M(\x; \myMat{D})\bigr]^\top,
  \qquad
  m_\ell(\x; \myMat{D}) = \prod_{j=1}^{n_x} x_j^{D_{j\ell}},
\label{eq:graphgrad_monomials}
\end{equation}
and the columns of $\myMat{D}$ enumerate all exponent vectors with total degree $\sum_{j=1}^{n_x} D_{j\ell} \leq d$, so that $M = \binom{n_x+d}{d}$. The column of zeros yields the constant monomial $m_\ell = 1$.
The LG-SSM in~\eqref{eq:ssm_state} is the special case of~\eqref{eq:graphgrad_model} with $d=1$ (up to the constant term), with $\myMat{C} = \myMat{A}$ and $\myVec{m}(\x_{t-1}; \myMat{D})=\x_{t-1}$. The directed graph $\mathcal{G}_{\x}$ is determined by the sparsity pattern of $\myMat{C}$: component $j$ directly influences component $i$ iff at least one monomial involving $x_j$ has a nonzero coefficient in the $i$-th row of $\myMat{C}$,
\begin{equation}
  j \to i \;\in\; \mathcal{G}_{\x}
  \quad\Longleftrightarrow\quad
  \exists\, \ell :
  D_{j\ell} > 0
  \text{ and }
  C_{i\ell} \neq 0.
\label{eq:graphgrad_graph}
\end{equation}
For example, the Euler--Maruyama discretization of the Lorenz~63 system is exactly of this form with $d=2$: each state equation combines linear terms with at most one bilinear term ($x_1x_3$ or $x_1x_2$), and the induced graph contains every directed edge except $3\to1$.

The MAP estimate solves
\begin{equation}
 \widehat{\myMat{C}}
 =
 \arg\min_{\myMat{C}}
 \Bigl(
   -\log \widehat{p(\y_{1:T} \mid \myMat{C})}
   + \lambda_C \|\myMat{C}\|_1
 \Bigr),
\label{eq:graphgrad_obj}
\end{equation}
where $\widehat{p(\y_{1:T} \mid \myMat{C})}$ is a differentiable approximation of the intractable likelihood obtained with a differentiable particle filter (DPF)~\cite{cox2025graphgrad}.

\begin{table*}[t]
\centering \caption{Comparison of graph-based SSM estimation methods.} \label{tab:methods} \smallskip
\begin{tabular}{lcccc}
\toprule
  & \textbf{GraphEM}~\cite{elvira2022graphical}
  & \textbf{DGLASSO}~\cite{chouzenoux2024sparse}
  & \textbf{GraphGrad}~\cite{cox2025graphgrad} \\
\midrule
Model class
  & LG-SSM
  & LG-SSM
  & General (nonlinear) SSM \\
Target parameter(s)
  & $\myMat{A}$
  & $\myMat{A}$, $\myMat{P}=\myMat{Q}^{-1}$
  & $\myMat{C}$ (polynomial coeff.) \\
Graph type
  & Directed
  & Directed + undirected
  & Directed \\
Inference alg.
  & Kalman filter/smoother
  & Kalman filter/smoother
  & Differentiable particle filter \\
Optimization alg.
  & EM + proximal M-step
  & Block alternating MM
  & Proximal gradient descent \\
Sparsity penalty
  & $\ell_1$ on $\myMat{A}$ 
  & $\ell_1$ on $\myMat{A}$ and $\myMat{P}$
  & $\ell_1$ on $\myMat{C}$ \\
\bottomrule
\end{tabular}
\end{table*}

\paragraph{Measurement-noise experiment}
{We compare observation-space VAR-GC against a penalized-likelihood LG-SSM estimator on {an LG-SSM with $n_x=5$ latent processes arranged in a directed chain $x_1\!\to\!x_2\!\to\!x_3\!\to\!x_4\!\to\!x_5$, with diagonal coefficients $A_{i,i}=0.2$ and off-diagonal coefficients $A_{i+1,i}=0.6$, all components observed ($\myMat{H}=\myMat{I}$),} with additive observation noise and varying noise level $\sigma_R$. When a VAR is fitted directly to noisy observations, the lagged regressors are themselves contaminated by noise (an errors-in-variables problem), which attenuates the estimated autoregressive coefficients toward zero as $\sigma_R$ increases. True directed effects therefore become progressively harder to detect. By contrast, the LG-SSM models the observation noise explicitly. Table~\ref{tab:exp_noise} shows the resulting failure mode: VAR-GC progressively loses true edges, while GraphEM~\cite{elvira2022graphical} retains all true edges across the tested noise levels.} However, with $T=500$ and the edge rule
$|\widehat{A}_{ij}|>0.05$, GraphEM also reports small spurious entries,
whose magnitude grows with $\sigma_R$, yielding lower precision and $F_1$
than VAR-GC. Thus, explicitly modeling measurement noise prevents false
negatives, but sparse graph recovery based on $\ell_1$ regularization from short noisy series still requires adequate thresholding.

\begin{table}[h]
\centering
\caption{\textbf{Measurement-noise experiment}: additive observation noise induces false negatives in VAR-based Granger causality estimation. Five-node chain, $T=500$, $\alpha=0.05$ with Bonferroni correction. The attenuated coefficient denotes the population OLS value associated with each true edge under measurement noise. Results are averaged over 500 independent Monte Carlo runs. Recall (R), precision (P) and $F_1$ are pooled over runs; a GraphEM edge is declared when $|\widehat{A}_{ij}|>0.05$, $i\neq j$.}
\label{tab:exp_noise}

\medskip
{\color{black}
\begin{tabular}{cccccccc}
\toprule
& & \multicolumn{3}{c}{VAR-GC} & \multicolumn{3}{c}{GraphEM} \\
\cmidrule(lr){3-5}\cmidrule(lr){6-8}
$\sigma_R$ & att.\ coeff. & R & P & $F_1$ & R & P & $F_1$ \\
\midrule
$0.0$ & $0.60$ & $1.00$ & $0.99$ & $1.00$ & $1.00$ & $0.60$ & $0.75$ \\
$0.5$ & $0.49$--$0.53$ & $1.00$ & $0.98$ & $0.99$ & $1.00$ & $0.49$ & $0.66$ \\
$1.0$ & $0.31$--$0.38$ & $1.00$ & $0.96$ & $0.98$ & $1.00$ & $0.37$ & $0.54$ \\
$1.5$ & $0.19$--$0.27$ & $0.82$ & $0.96$ & $0.88$ & $1.00$ & $0.31$ & $0.47$ \\
$2.0$ & $0.13$--$0.19$ & $0.38$ & $0.94$ & $0.54$ & $1.00$ & $0.31$ & $0.47$ \\
\bottomrule
\end{tabular}}
\end{table}

\subsection*{Identifiability}
\label{sec:identifiability}

For a generic LG-SSM, different parameter triples $(\myMat{A}, \myMat{H}, \myMat{Q})$ can generate the same marginal distribution over $\y_{1:T}$: for any invertible $\myMat{T} \in \mathbb{R}^{n_x \times n_x}$, the transformed system $(\myMat{T}\myMat{A}\myMat{T}^{-1},\; \myMat{H}\myMat{T}^{-1},\; \myMat{T}\myMat{Q}\myMat{T}^\top)$, with initial distribution $\mathcal{N}(\myMat{T}\myVec{\mu}_0,\myMat{T}\myMat{\Sigma}_0\myMat{T}^\top)$, produces identical observations. This \emph{similarity class} has $n_x^2$ degrees of freedom, and the transformation generally changes the sparsity pattern of $\myMat{A}$~\cite{ljung1999system}.

\paragraph{Observation geometry}
When $\myMat{H}$ is known and full column rank with $n_y \geq n_x$, the constraint $\myMat{H}\myMat{T}^{-1}=\myMat{H}$ forces $\myMat{T}=\myMat{I}$, eliminating the similarity ambiguity (including the common setting $n_x=n_y$, $\myMat{H}=\myMat{I}$); identifiability of $\myMat{A}$ then depends on the usual statistical conditions, such as sufficient excitation of the state dynamics. When $n_y < n_x$, $\myMat{H}$ cannot have full column rank, leaving a non-trivial family of $\myMat{T}$ with $\myMat{H}\myMat{T}^{-1}=\myMat{H}$, and hence of equivalent models, and additional restrictions are then required~\cite{ljung1999system}.

\paragraph{Structural restrictions}
When structural information about the latent system is available, it can be encoded as a hard restriction on the parameter space, which rules out some of the equivalent models:
\begin{itemize}[leftmargin=*]
  \item \emph{Structural masks on $\myMat{A}$}: imposing $A_{ij}=0$ for physically impossible connections rules out every equivalent model whose transition matrix $\myMat{T}\myMat{A}\myMat{T}^{-1}$ does not keep these zeros, as with the spatial neighborhood constraints of~\cite{luengo2018hierarchical}. 

  \item \emph{Independent state noise}: a diagonal $\myMat{Q}$ removes only part of the ambiguity: with $\myMat{Q}$ fixed, $\myMat{T}\myMat{Q}\myMat{T}^\top=\myMat{Q}$ still allows the continuous family $\myMat{T}=\myMat{Q}^{1/2}\myMat{O}\myMat{Q}^{-1/2}$, $\myMat{O}$ orthogonal. Latent rotations remain, and reducing the ambiguity to signed permutations requires breaking them, e.g., with structural masks or non-Gaussian noise.

\item \emph{Canonical parametrizations}:  
canonical forms, such as the observable echelon form, fix one representative per similarity class by imposing a known pattern of zeros and ones on $\myMat{H}$ and part of $\myMat{A}$~\cite{bryutkin2025canonical,ljung1999system}. Under standard regularity conditions this makes the model identifiable, but the zeros are imposed by the parametrization, so the sparsity pattern of the canonical $\myMat{A}$ cannot be read as $\mathcal{G}_{\x}$.
\end{itemize}

\paragraph{Sparse representatives when structure is unknown}

When such restrictions are unavailable, the natural fallback is to estimate $\myMat{A}$ under a sparsity-inducing penalty, as reviewed above. The likelihood remains constant along the similarity class~\cite{ljung1999system}, so the model is still not identifiable~\cite{gustafson2009limits}; the penalty only selects, among equally good fits, one with small $\ell_1$ norm. Permutations and diagonal scalings of the latent state preserve the sparsity pattern of $\myMat{A}$, so a sparse representative can be unique at most up to these transformations; fixing them requires information that maps latent components to physical variables, such as a known $\myMat{H}$ or a structural mask.
Beyond that, we are not aware of a general uniqueness result: support-recovery guarantees for sparse regression~\cite{zhao2006model,wainwright2009sharp} apply once the model is identified, not to the selection within a class.

\paragraph{What can be read from the recovered graph}
The eigenvalues of $\myMat{A}$ and the spectral density of $\y_t$
(hence the observation-space Granger graph) are invariant across the
similarity class. However, the sparsity pattern of $\myMat{A}$ is not,
so reading $j\to i$ as an interaction between physical variables requires
anchoring the latent coordinates, e.g., through a known $\myMat{H}$ or a
structural mask. The recovered graph $\mathcal{G}_{\x}$ therefore says more than an observation-space Granger graph, because it describes the latent dynamics and can avoid the failure modes of the previous section under correct model specification. It says less than a causal graph, which additionally requires reading the transition equations as structural mechanisms, as discussed next.

\paragraph{Nonlinear constraints}
\label{sec:exp_constraints}

We illustrate how structural constraints improve graph recovery under nonlinear dynamics and model misspecification. Following Luengo et al.~\cite{luengo2018hierarchical}, a 10-pole circular catheter yielding nine bipolar channels records synthetic intracardiac electrograms generated by coupled FitzHugh--Nagumo oscillators. The dynamics are nonlinear, so neither the VAR nor the LG-SSM assumptions hold exactly. The ground-truth propagation graph is known, and physiology provides an additional constraint: electrical wavefronts propagate locally, so electrode $i$ can directly influence only spatial neighbors ($d_{\max}=1$). The dominant factor is therefore not whether inference is performed in observation or latent space, but whether the admissible graph class is structurally constrained.

We compare pairwise GC, full-conditional GC, PCMCI~\cite{runge2019detecting}, LS-CaRe~\cite{luengo2018hierarchical}, and GraphEM with the same spatial mask imposed after each M-step. {For PCMCI we run two variants as an ablation: a linear partial-correlation test (PCMCI$_{\rm lin}$) and a nonlinear $k$-NN conditional mutual information test (PCMCI$_{\rm nl}$). The purpose is to separate the contribution of the conditional-independence test from that of the structural constraint.} Table~\ref{tab:exp_fhn} shows that the best method in every row uses the locality mask. {The two PCMCI variants obtain similar $F_1$ scores, so the choice of test does not explain their low accuracy.} LS-CaRe performs best for single-rotor and circular propagation, while GraphEM performs best for flat wavefronts. GraphEM is less robust in the single-rotor regime.

\begin{table}[t]
\centering
\footnotesize
\caption{\textbf{Nonlinear-constraints experiment}: $F_1$ scores on FitzHugh--Nagumo electrogram data~\cite{luengo2018hierarchical}. The best-performing method in every setting uses the spatial locality constraint $d_{\max}=1$ (LS-CaRe, GraphEM). {PCMCI$_{\rm lin}$: linear partial correlation and PCMCI$_{\rm nl}$: nonlinear.} Bold: best method per row.}
\label{tab:exp_fhn}
\setlength{\tabcolsep}{3.5pt}
\begin{tabular}{llcccccc}
\toprule
& & \multicolumn{4}{c}{\textit{No constraint}}
& \multicolumn{2}{c}{\textit{$d_{\max}=1$}} \\
\cmidrule(lr){3-6}\cmidrule(lr){7-8}
\textbf{Pattern} & \textbf{SNR}
& Pair & Full & {PCMCI$_{\rm lin}$} & {PCMCI$_{\rm nl}$}
& LS-CaRe & GraphEM \\
\midrule
Single
& 10\,dB & 0.63 & 0.50 & 0.13 & {0.16} & \textbf{1.00} & 0.64 \\
& 20\,dB & 0.58 & 0.75 & 0.18 & {0.13} & \textbf{1.00} & 0.50 \\
& 40\,dB & 0.44 & 0.62 & 0.14 & {0.12} & \textbf{0.90} & 0.53 \\
\midrule
Circular
& 10\,dB & 0.09 & 0.00 & 0.05 & {0.18} & \textbf{0.71} & 0.60 \\
& 20\,dB & 0.17 & 0.13 & 0.18 & {0.15} & \textbf{0.88} & 0.57 \\
& 40\,dB & 0.23 & 0.21 & 0.16 & {0.19} & \textbf{0.82} & 0.67 \\
\midrule
Flat
& 10\,dB & 0.56 & 0.17 & 0.22 & {0.20} & 0.71 & \textbf{0.80} \\
& 20\,dB & 0.60 & 0.59 & 0.24 & {0.19} & 0.50 & \textbf{0.63} \\
& 40\,dB & 0.47 & 0.58 & 0.16 & {0.12} & 0.56 & \textbf{0.63} \\
\bottomrule
\end{tabular}
\end{table}

\begin{figure*}[t]
\centering
\begin{tikzpicture}[
    box_base/.style={rectangle, rounded corners=4pt, draw, line width=0.8pt,
        text width=4.2cm, align=center,
        minimum height=2.5cm, font=\small, inner sep=1pt},
    scm/.style={box_base, fill=blue!8, draw=blue!55},
    sdcm/.style={box_base, fill=violet!8, draw=violet!55},
    ssm/.style={box_base, fill=green!8, draw=green!55},
    arr/.style={-{Stealth[length=3mm]}, line width=1.0pt, gray!50!black},
    arr_label/.style={font=\footnotesize\sffamily, align=center}
]
\node[scm]  (a) at (-6.7, 0) {%
    \textbf{SCM (Pearl)} \\[3pt]
    $X_i = f_i(X_{\mathrm{Pa}(i)},\, U_i)$ \\[3pt]
    \footnotesize Static; acyclic graph; intervention $\mathrm{do}(X_i\!=\!x)$ replaces equation~$i$.};
\node[sdcm] (b) at (-0.5, 0) {%
    \textbf{SDCM (Bongers)} \\[3pt]
    $\dot{X}_i(t) = g_i\!\bigl(X_{\mathrm{Pa}(i)}(t),\, U_i(t)\bigr)$ \\[3pt]
    \footnotesize Dynamical; feedback via time unrolling; modularity per equation.};
\node[ssm]  (c) at ( 5.7, 0) {%
    \textbf{Discrete-time SSM} \\[3pt]
    $\x_t = f(\x_{t-1}) + \vw_t$ \\[3pt]
    \footnotesize Linear $f$: LG-SSM. Polynomial $f$: GraphGrad. Parents encoded by sparsity of $f$.};
\draw[arr] (a.east) -- (b.west);
\node[arr_label, above=1pt] at ($(a.east)!0.5!(b.west)$) {$+$ time,\\ $+$ feedback};
\draw[arr] (b.east) -- (c.west);
\node[arr_label, above=1pt] at ($(b.east)!0.5!(c.west)$) {discretize};
\end{tikzpicture}
\caption{Conceptual relation between SCMs, SDCMs, and state-space models. SCMs encode static causal mechanisms; SDCMs extend them to dynamical systems with feedback and interventions on trajectories; discretizing an SDCM yields the transition equations of an SSM, which can also be read structurally on its own.}
\label{fig:sidebar_sdcm}
\end{figure*}
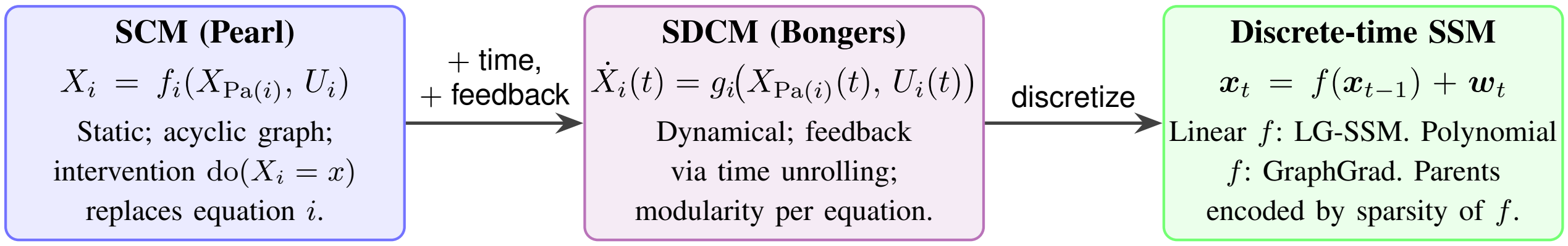

\subsection*{\texorpdfstring{Causal interpretation of the latent graph}{Causal interpretation of the latent graph}}
\labelname{Causal interpretation of the latent graph}\label{sec:sdcm_causal}

The methods presented above recover directed structure from latent dynamics: $j \to i$ holds iff $A_{ij} \neq 0$ in the linear case~\eqref{eq:latent_graph}, or iff $x_{t-1,j}$ appears in $f_i(\x_{t-1})$ in the polynomial case~\eqref{eq:graphgrad_graph}.
A discrete-time SSM can be given a structural causal interpretation directly, by treating each componentwise transition equation as an autonomous mechanism that an intervention replaces while the others remain unchanged. In discrete time, a hard intervention on component $i$ means externally fixing $x_{t,i}=c$ while all other components evolve under their own structural equations. Structural dynamical causal models (SDCMs)~\cite{bongers2018causal} provide the corresponding formalism for continuous-time dynamical systems, where interventions act on differential equations. Two further questions are separate from this interpretation: whether the recovered graph is the graph of the system (identifiability), and, when the discrete-time model is meant to represent an underlying continuous-time system, how its transition graph relates to the instantaneous couplings of that system (discretization). Figure~\ref{fig:sidebar_sdcm} summarizes the relation between SCMs, SDCMs, and SSMs.

\paragraph{From SCMs to SDCMs}
A structural causal model (SCM)~\cite{pearl2009causality} assigns to each variable $X_i$ a structural equation $X_i = f_i(\mathrm{Pa}(X_i), U_i)$ with exogenous noise $U_i$; the causal graph is defined by the parent sets $\mathrm{Pa}(X_i)$. SDCMs extend this idea to dynamical systems: each component $i$ of the state is governed by a \emph{dynamic structural equation},
\begin{equation}
  \dot{X}_i(t) = g_i\!\bigl(X_{\mathrm{Pa}(i)}(t),\, U_i(t)\bigr),
\label{eq:sdcm}
\end{equation}
where $X_{\mathrm{Pa}(i)}$ denotes the components driving the $i$-th mechanism. The key property is \emph{modularity}: each equation represents an autonomous mechanism that can be intervened upon independently of the others. Unlike static SCMs, SDCMs naturally accommodate feedback: cyclic mechanisms in the rolled graph (e.g., $x_1 \to x_2 \to x_3 \to x_1$) are allowed because the time-unrolled graph $\cdots \to x_{t,1} \to x_{t+1,2} \to x_{t+2,3} \to x_{t+3,1} \to \cdots$ remains acyclic by temporal order.

\paragraph{Continuous and discrete time}
Consider a linear SDCM
$\dot{\x}(t)=\myMat{G}\x(t)+\myVec{u}(t)$,
where $\myMat{G}$ is the drift matrix and $\myVec{u}(t)$ is a continuous-time stochastic forcing process. An Euler step gives $\myMat{A}=\myMat{I}+\Delta t\,\myMat{G}$, whose off-diagonal zero pattern is exactly that of $\myMat{G}$. Exact sampling instead gives $\myMat{A}=\exp(\myMat{G}\Delta t)=\myMat{I}+\Delta t\,\myMat{G}+\tfrac{1}{2}\Delta t^2\myMat{G}^2+\cdots$, where directed paths of length $k$ in $\myMat{G}$ contribute through the term $\Delta t^k\myMat{G}^k/k!$; an entry $A_{ij}$ with $G_{ij}=0$ whose shortest path from $j$ to $i$ has length $k$ is therefore generally nonzero, with leading order $\Delta t^k$. Likewise, a diagonal instantaneous noise covariance generally yields a non-diagonal discrete-time noise covariance, with off-diagonal terms appearing at order $\Delta t^2$. The graph of a sampled transition therefore describes dependencies accumulated over a sampling interval and need not coincide with the graph of instantaneous couplings; this can be seen as the continuous-time counterpart of subsampling a VAR process. If the drift of the cardiovascular example of Figure~\ref{fig:running_example} has the fork structure $x_2\leftarrow x_1\rightarrow x_3$, there is no directed path between any pair of distinct components that is not already connected by an edge, so exact sampling adds no off-diagonal edge; if the drift of the five-node chain is itself a chain, exact sampling adds, e.g., the edge $1\to3$ through the term $\tfrac{1}{2}\Delta t^2\,(\myMat{G}^2)_{31}$. For the polynomial model, an Euler step likewise gives a coefficient matrix $\myMat{C}$ with the parent structure of the continuous-time mechanism.

\paragraph{Requirements for a causal interpretation of the transition graph}
Interpreting the directed transition graph causally requires additional assumptions{:}
\begin{enumerate}[leftmargin=*]
 \item \emph{Continuous-time correspondence.}
If the SSM is obtained by sampling an underlying continuous-time
system, the graph of the exact discrete-time transition (in the linear case, $\myMat{A}=\exp(\myMat{G}\Delta t)$) may contain additional edges arising
from indirect paths in the instantaneous causal graph of the continuous-time system, as discussed above.

  \item \emph{Independent noise components.}
  The transition matrix $\myMat{A}$ captures only directed dependencies
  across time. If the innovations of different state components are dependent,
  additional contemporaneous dependencies are present that are not represented
  in the transition graph. Thus, interpreting the graph of $\myMat{A}$ as the
  full causal structure requires independent innovations; otherwise, the
  innovation dependence must also be represented, as in models that jointly
  estimate the transition and innovation graphs. If $\myMat{Q}$ is block diagonal, innovations are independent across blocks but may remain dependent within each block; the transition graph alone therefore does not represent the dependencies within those blocks.
  
  \item \emph{Causal sufficiency.}  The latent state $\x_t$ must contain all causally relevant processes. {Otherwise the transition parameters absorb indirect pathways and mediated effects. The  hidden-intermediate-node experiment  illustrates this directly:  omitting $x_3$ leads VAR-GC to infer spurious direct edges, and with $n_x=4$ GraphEM cannot recover the missing node. The choice of latent dimension $n_x$ and observation operator $\myMat{H}$ is therefore itself a causal modeling assumption: it defines which processes are treated as part of the causal system.}
\end{enumerate}

These requirements concern the causal interpretation of the transition
graph and are distinct from identifiability, which determines whether the
transition graph can be recovered from the observations.

\section*{\texorpdfstring{Practical guidelines and illustrative examples}{Practical guidelines and illustrative examples}}
\label{sec:practical}

\subsection*{Choosing a method: decision guide}
\label{sec:decision}

The methods surveyed in this tutorial span a broad spectrum of assumptions and complexity. The decision logic can be summarized as two sequential questions: first, whether observation-space assumptions are credible (i.e., no preprocessing-induced VARMA structure and no plausible latent confounders); second, given the chosen space, whether dimensionality (observation branch) or nonlinearity (latent branch) is the dominant constraint.  

\paragraph{Observation-space assumptions}
Observation-space methods require that the observed process admits a VAR representation and that no relevant driver is hidden. Both assumptions fail in common situations: filtering, subsampling, or averaging generally produce a VARMA process, and unobserved drivers cannot be corrected by observation-space estimators that assume causal sufficiency; methods that allow latent confounders, such as tsFCI~\cite{entner2010causal} and LPCMCI~\cite{gerhardus2020high}, can flag them but do not recover the latent structure. When latent confounders are plausible, SSMs offer a way to model the latent processes explicitly. Within the observation-space setting, sparsity penalties (Lasso-VAR) and constraint-based discovery (PCMCI) are useful when $n_y$ is large relative to $T$ or when the series are strongly autocorrelated; classical $F$-tests on OLS-VAR remain a reasonable baseline otherwise.

\paragraph{Identifiability and causal interpretation}

Two questions should be kept apart. Identifiability asks whether the latent graph can be recovered from the data; it usually requires a known observation model or structural constraints, since sparsity penalties alone only select one representative among equivalent models. Causal interpretation asks what the graph means: it can be read causally when the transition equations are taken as structural mechanisms with their own drivers and no relevant process is omitted; otherwise it describes predictive structure in latent space.
 
\paragraph{Computational complexity and scalability}
Lasso-VAR estimates $n_y^2p$ coefficients, and PCMCI has a complexity polynomial in $n_y$ and the maximum lag~\cite{runge2019detecting}. In latent-state methods, each Kalman filter pass costs $\mathcal{O}(T(n_x^3+n_y^3))$~\cite{cox2023sparse}: GraphEM runs a Kalman filter and smoother at each EM iteration, DGLASSO twice per iteration \cite{chouzenoux2024sparse}; both run inner solvers which may dominate the complexity. SpaRJ runs one Kalman filter pass per MCMC iteration, which accounts for most of its run time~\cite{cox2023sparse}. GraphGrad instead runs a differentiable particle filter at each gradient step, each pass being linear in $T$ and in the number of particles~\cite{cox2025graphgrad}.

\paragraph{Software} The MVGC toolbox~\cite{barnett2014mvgc} implements time- and frequency-domain Granger causality for VAR models, while a state-space formulation is developed in~\cite{barnett2015granger}; tigramite implements PCMCI and its nonlinear extensions~\cite{runge2019detecting}. Reference implementations for the latent-state methods are available in the authors' repositories.

\subsection*{\texorpdfstring{Hidden intermediate node experiment}{Hidden intermediate node experiment}}
\labelname{Hidden intermediate node experiment}\label{sec:exp_confounding}

{We use the same five-node chain as in the measurement-noise experiment ($x_1\!\to\!x_2\!\to\!x_3\!\to\!x_4\!\to\!x_5$, $A_{i,i}=0.2$, $A_{i+1,i}=0.6$)}, but now observe only $x_1,x_2,x_4,x_5$. Observation noise is Gaussian with covariance $0.25\,\myMat{I}$, and we generate $T=50{,}000$ samples.
The sub-chain $x_2\to x_3\to x_4$ is the hidden-mediator case, the counterpart of the hidden common cause in Fig.~\ref{fig:running_example}: the effect of $x_2$ on $x_4$ is real, but observation-space methods report it as a direct link that bypasses the missing node.

\noindent\textbf{Observation-space methods.} We apply full-conditional Granger causality with $p=5$ lags and Bonferroni correction at $\alpha=0.05$ (GCCA toolbox~\cite{seth2010matlab}), and PCMCI with linear partial-correlation tests~\cite{runge2019detecting}. Both recover the two true observable edges ($x_1\!\to\!x_2$, $x_4\!\to\!x_5$), but also infer three spurious direct edges (red in Fig.~\ref{fig:exp_confounding}(b)). These are predictive dependencies induced by the hidden mediator, not finite-sample artifacts: the $F$-statistic of $x_2\!\to\!x_4$ grows from $19.1$ at $T=2{,}000$ to $799.8$ at $T=50{,}000$.

\noindent\textbf{State-space approach.} GraphEM~\cite{elvira2022graphical} with $n_x=5$, known $\myMat{H}$, and $\ell_1$+stability regularization ($\lambda_A=0.5$) recovers the full latent chain including $x_3$ (Fig.~\ref{fig:exp_confounding}(c)). Declaring a GraphEM edge $j\to i$ when $|\widehat{A}_{ij}|>0.05$, $i\neq j$ (the rule used in all experiments), GraphEM attains precision, recall and $F_1$ equal to $1$, whereas Granger causality, scored against the direct observable edges $x_1\to x_2$ and $x_4\to x_5$, has recall $1$ but precision $0.40$ ($F_1=0.57$). This requires specifying the latent dimension and the observation operator correctly; with $n_x=4$, GraphEM cannot recover the missing latent node.

{The two approaches target different graphical objects. Observation-space Granger methods estimate the predictive graph of $\y_t$, while the SSM formulation targets the latent transition graph $\mathcal{G}_{\x}$.} All experiments in this tutorial are illustrations of specific mechanisms with known ground truth, not a benchmark. They do not address misspecified $n_x$ or $\myMat{H}$, the choice of regularization parameters without ground truth, scalability, or real data; for comparisons we refer to the original publications~\cite{elvira2022graphical,chouzenoux2024sparse,cox2025graphgrad,runge2019detecting,luengo2018hierarchical}.

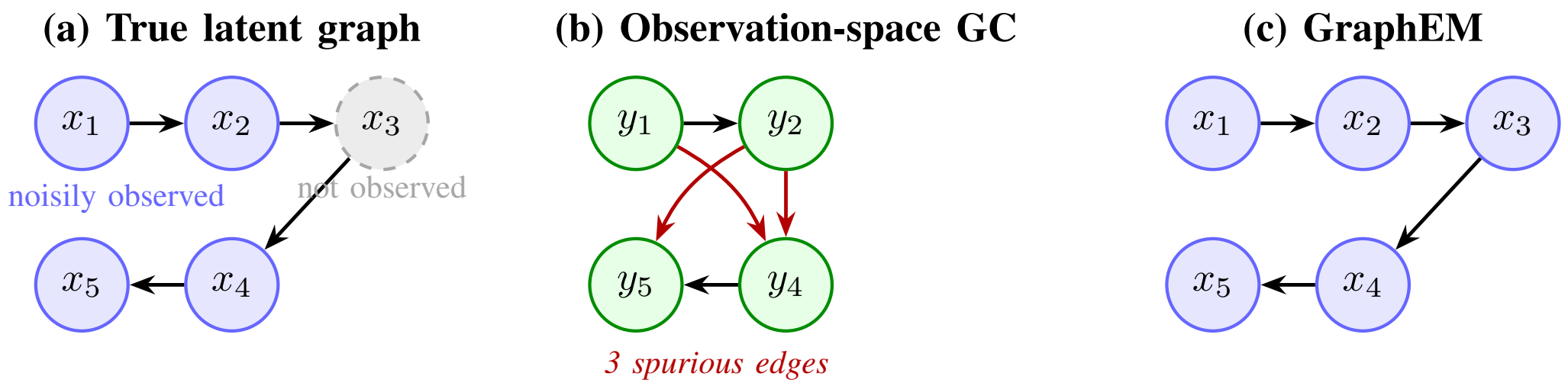
\begin{figure}[t]
\centering
\begin{tikzpicture}[
    latentobs/.style={
        circle,
        draw=blue!60,
        fill=blue!10,
        thick,
        minimum size=8mm,
        font=\small
    },
    latenthidden/.style={
        circle,
        draw=gray!70,
        fill=gray!15,
        thick,
        dashed,
        minimum size=8mm,
        font=\small
    },
    obsnode/.style={
        circle,
        draw=green!55!black,
        fill=green!10,
        thick,
        minimum size=8mm,
        font=\small
    },
    recovered/.style={
        circle,
        draw=green!60!black,
        fill=green!10,
        thick,
        dashed,
        minimum size=8mm,
        font=\small
    },
    true_edge/.style={
        -{Stealth[length=2.2mm]},
        line width=0.9pt,
        black
    },
    obs_edge/.style={
        -{Stealth[length=2.2mm]},
        line width=0.9pt,
        red!70!black
    },
    panel/.style={
        font=\small\bfseries
    }
]

\begin{scope}[shift={(0,0)}]

\node[panel] at (1.3,2.8)
{(a) True latent graph};

\node[latentobs]    (a1) at (0,2)   {$x_1$};
\node[latentobs]    (a2) at (1.3,2) {$x_2$};
\node[latenthidden] (a3) at (2.6,2) {$x_3$};
\node[latentobs]    (a4) at (1.3,0.6) {$x_4$};
\node[latentobs]    (a5) at (0,0.6) {$x_5$};

\draw[true_edge] (a1) -- (a2);
\draw[true_edge] (a2) -- (a3);
\draw[true_edge] (a3) -- (a4);
\draw[true_edge] (a4) -- (a5);

\node[font=\scriptsize,blue!60]
at (0.3,1.35) {noisily observed};

\node[font=\scriptsize,gray!70]
at (2.6,1.45) {not observed};

\end{scope}

\begin{scope}[shift={(4.8,0)}]

\node[panel] at (1.3,2.8)
{(b) Observation-space GC};

\node[obsnode] (b1) at (0,2)   {$y_1$};
\node[obsnode] (b2) at (1.3,2) {$y_2$};
\node[obsnode] (b4) at (1.3,0.6) {$y_4$};
\node[obsnode] (b5) at (0,0.6) {$y_5$};

\draw[true_edge] (b1) -- (b2);
\draw[true_edge] (b4) -- (b5);

\draw[obs_edge] (b2) -- (b4);

\draw[obs_edge,bend left=18]
(b1) to (b4);

\draw[obs_edge,bend right=18]
(b2) to (b5);

\node[font=\scriptsize\itshape,red!70!black]
at (0.7,-0.1)
{3 spurious edges};

\end{scope}

\begin{scope}[shift={(9.8,0)}]

\node[panel] at (1.3,2.8)
{(c) GraphEM};

\node[latentobs] (c1) at (0,2)   {$x_1$};
\node[latentobs] (c2) at (1.3,2) {$x_2$};
\node[latentobs] (c3) at (2.6,2) {$x_3$};
\node[latentobs] (c4) at (1.3,0.6) {$x_4$};
\node[latentobs] (c5) at (0,0.6) {$x_5$};

\draw[true_edge] (c1) -- (c2);
\draw[true_edge] (c2) -- (c3);
\draw[true_edge] (c3) -- (c4);
\draw[true_edge] (c4) -- (c5);

\end{scope}

\end{tikzpicture}

\caption{
\textbf{Hidden intermediate node experiment}.
(a) True latent graph $\mathcal{G}_{\x}$.
Blue nodes are latent components indirectly measured through noisy observations;
$x_3$ (gray dashed) is completely hidden.
(b) Observation-space Granger causality applied to the observed processes
$\y_t=(y_1,y_2,y_4,y_5)^\top$.
Besides the true observable edges, three spurious direct edges appear
(red) due to the hidden mediator $x_3$.
(c) GraphEM reconstructs the latent transition graph using the same $\y_t$ and recovers
the true $\mathcal{G}_{\x}$.
Results are from a single run at $T=50{,}000$.}
\label{fig:exp_confounding}

\end{figure}

\subsection*{\texorpdfstring{Related approaches and their place on the causal ladder}{Related approaches and their place on the causal ladder}}
\label{sec:related}

Graph signal processing (GSP) usually processes signals on a known graph, e.g., graph VARMA models for forecasting~\cite{isufi2019forecasting}; related work on network topology inference instead learns the graph from the signals~\cite{mateos2019connecting,dong2019learning}, with variants for directed~\cite{Mei2017,marques2020signal,cui2024topology} and time-varying~\cite{zaman2021online} graphs and for nonlinear dependencies~\cite{Giannakis2018}. Convergent cross mapping and directed-information tests address causal discovery in nonlinear and discrete time series~\cite{butler2023causal,theocharous2024temporally}. Most of these methods place the graph on the observed signals, i.e., on the first step of the ladder in Fig.~\ref{fig:causal_ladder}. Their connection with latent state-space models is beyond the scope of this tutorial.

Other families go beyond a graph among the observed signals, but their nodes need not be physical variables. Koopman methods~\cite{brunton2022modern} map the state through a set of functions in which the nonlinear dynamics become approximately linear. The resulting transition graph is therefore defined over these functions, whereas in the SSMs of this tutorial the nodes are the state components. Dynamic causal modeling~\cite{friston2003dynamic,friston2014granger} also models latent dynamics and the measurement process explicitly, but in its classical form selects among a few hypothesized graphs by Bayesian model evidence instead of learning a sparse graph. Neural Granger methods~\cite{tank2021neural}, in contrast, remain in the observation space, whereas temporal causal representation learning recovers latent causal processes, identifiable up to permutation and componentwise transformations under nonstationarity or non-Gaussianity assumptions~\cite{yao2022temporally}.

\section*{Conclusion and outlook}
\label{sec:conclusion}

This tutorial has examined a progression from classical Granger causality on observed time series to causal structure in latent dynamical systems. A central message is that these approaches do not estimate the same graphical object: observation-space methods target predictive structure in $\y_t$, while state-space formulations target the transition structure of an assumed model with latent structure, which in many applications is closer to the physical mechanisms that generate the data. Each step along this progression requires stronger assumptions: separating latent dynamics from observation effects, identifying the latent representation, and reading the transition equations as structural mechanisms. When the latent model samples an underlying continuous-time system, the recovered graph describes dependencies accumulated over one sampling interval, which may include indirect paths of the instantaneous causal graph. Specifying the latent representation is itself a causal modeling assumption.

An important open problem is the extension of the last two steps of the causal ladder to nonlinear systems: while identifiability is comparatively well characterized in linear Gaussian SSMs, it is unclear under which conditions nonlinear latent dynamical systems admit identifiable and causally interpretable representations.

\bibliographystyle{IEEEbib} %\bibliography{IEEEabrv,references}

\section*{Biography}
\vspace{-30mm}
\begin{IEEEbiographynophoto}{V\'ictor Elvira}
(\emph{Senior Member, IEEE}) is a Distinguished Research Professor at the Basque Centre for Applied Mathematics (BCAM), Bilbao, Spain, and a Professor with a Personal Chair in Statistics
and Data Science at the School of Mathematics, University of Edinburgh. He received his
Ph.D.\ in statistical signal processing from the University of Cantabria (Spain) in 2011.
His research interests are in statistical signal processing, computational statistics, and
probabilistic machine learning, with a focus on state-space models, Monte Carlo methods,
and graph-based formulations for time-series data. He is an ELLIS Fellow, Fulbright
Fellow, Marie Curie Fellow, Leverhulme Fellow, Alan Turing Fellow, and recipient of the
2024 EURASIP Early Career Award. He is the Principal Investigator of an ARIA-funded
project (2025--2028) on probabilistic forecasting of climate tipping points.
\end{IEEEbiographynophoto}

\end{document}